\pdfoutput=1
\documentclass[conference]{IEEEtran}
\usepackage{times}

\usepackage[numbers]{natbib}
\usepackage{multicol}
\usepackage[bookmarks=true]{hyperref}

\usepackage{color}
\usepackage[svgnames]{xcolor}
\usepackage{graphicx}       
\usepackage{subcaption}     

\usepackage{booktabs}       
\usepackage{multirow}       
\usepackage{amsfonts}       
\usepackage{amsmath}        
\usepackage{nicefrac}       
\usepackage{microtype}      
\let\labelindent\relax      
\usepackage{enumitem}       
\usepackage{siunitx}        

\usepackage{algorithm} 
\usepackage{algorithmic} 

\colorlet{Mycolor1}{green!10!orange!90!}

\newcommand{\Skip}[1]{}

\newcommand{\ie}{i.e.,\ }
\newcommand{\eg}{e.g.,\ }

\newcommand{\mysecref}[1]{Section~\ref{#1}}
\newcommand{\myfigref}[1]{Fig.~\ref{#1}}
\newcommand{\mytbref}[1]{Table~\ref{#1}}
\newcommand{\myeqref}[1]{Equation~\ref{#1}}
\newcommand{\myalgref}[1]{Algorithm~\ref{#1}}

\newcommand{\method}{{IDAPT}}
\newcommand{\fullmethod}{{Iterative Domain Alignment for Policy Transfer}}

\newcommand\blfootnote[1]{%
  \begingroup
  \renewcommand\thefootnote{}\footnote{#1}%
  \addtocounter{footnote}{-1}%
  \endgroup
}

\begin{document}


\title{
Policy Transfer across Visual and Dynamics\\Domain Gaps via Iterative Grounding
}



\author{\authorblockN{Grace Zhang}
\authorblockA{USC \\ \texttt{gracez@usc.edu}}
\and
\authorblockN{Linghan Zhong}
\authorblockA{USC \\ \texttt{linghanz@usc.edu}}
\and
\authorblockN{Youngwoon Lee$^\ast$}
\authorblockA{USC \\ NAVER AI Lab \\ \texttt{lee504@usc.edu}}
\and
\authorblockN{Joseph J. Lim}
\authorblockA{USC \\ NAVER AI Lab$^\dagger$ \\ \texttt{limjj@usc.edu}}
}



%

\maketitle

\begin{abstract}
The ability to transfer a policy from one environment to another is a promising avenue for efficient robot learning in realistic settings where task supervision is not available. This can allow us to take advantage of environments well suited for training, such as simulators or laboratories, to learn a policy for a real robot in a home or office. To succeed, such policy transfer must overcome both the visual domain gap (e.g. different illumination or background) and the dynamics domain gap (e.g. different robot calibration or modelling error) between source and target environments. However, prior policy transfer approaches either cannot handle a large domain gap or can only address one type of domain gap at a time. In this paper, we propose a novel policy transfer method with iterative ``environment grounding'', IDAPT, that alternates between (1) directly minimizing both visual and dynamics domain gaps by grounding the source environment in the target environment domains, and (2) training a policy on the grounded source environment. This iterative training progressively aligns the domains between the two environments and adapts the policy to the target environment. Once trained, the policy can be directly executed on the target environment. The empirical results on locomotion and robotic manipulation tasks demonstrate that our approach can effectively transfer a policy across visual and dynamics domain gaps with minimal supervision and interaction with the target environment. Videos and code are available at \url{https://clvrai.com/idapt}.

\end{abstract}
\blfootnote{$^\ast$Work done during an internship at NAVER AI Lab.}
\blfootnote{$^\dagger$AI Advisor.}

\IEEEpeerreviewmaketitle

\section{Introduction}

\begin{figure}[t]
    \centering
    \includegraphics[width=0.9\linewidth]{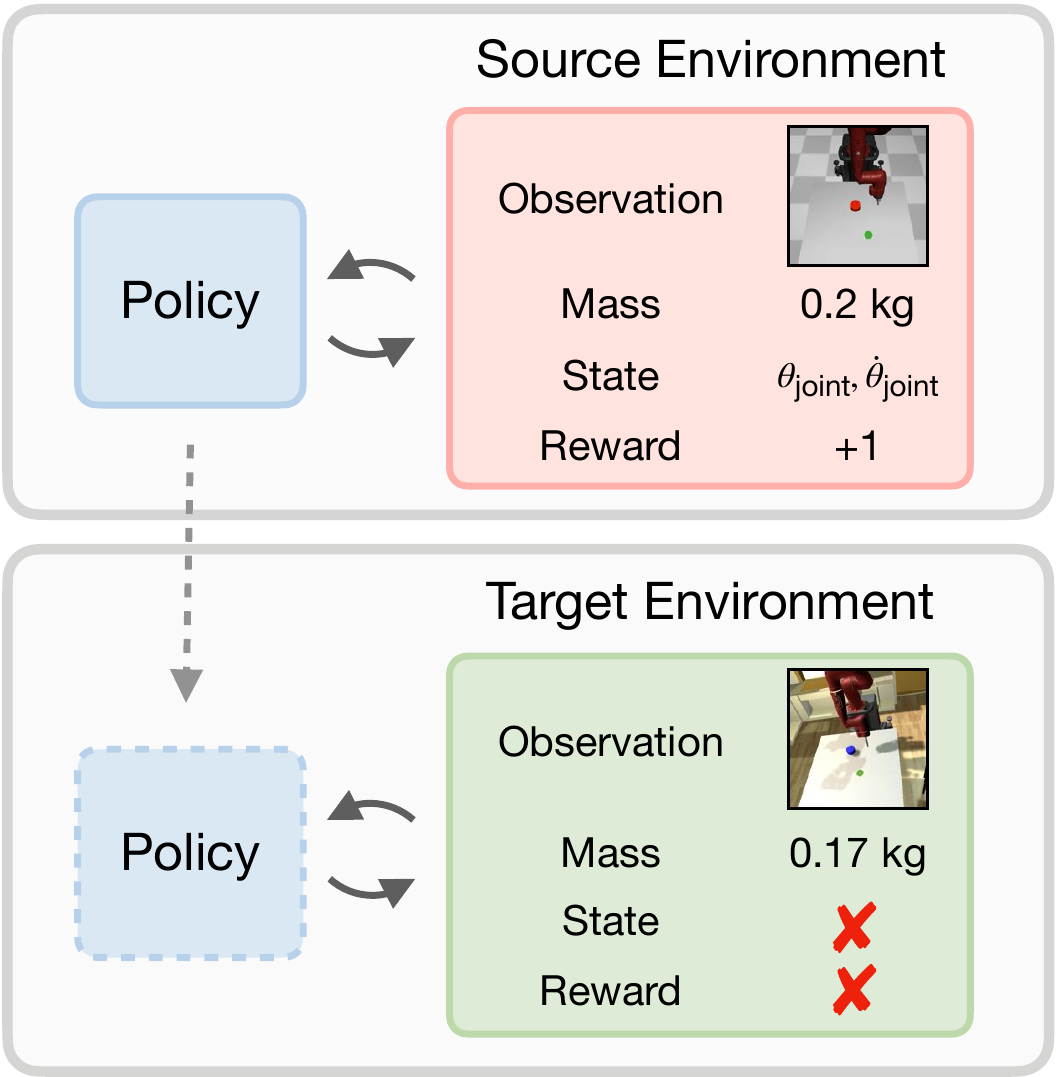}
    \caption{
        We aim to transfer a policy trained in the source environment (top) which has full access to observation, state, and reward, to the target domain (bottom) with no access to state and reward. However, transferring a policy from one environment to another is challenging due to visual differences (e.g. lighting, viewpoint, clutter, background) and physical differences (e.g. friction, mass, robot calibration).      
    }
    \label{fig:teaser}
\end{figure}

Deep reinforcement learning (RL) presents a promising framework for learning impressive robot behaviors~\citep{levine2016end, suarez2016framework, rajeswaran2018learning, jain2019learning, lee2021ikea}. Yet, performing RL directly on physical robots in a home or an office is impractical due to the lack of training supervision (e.g. reward function, state information) and the high cost of data collection, which holds true in most robot learning scenarios. 
One practical solution is \textit{policy transfer}, which first trains a policy under a more controllable environment (often a simulator) with cheaper data collection and easier access to reward and state information, and then deploys this well-trained policy to the target environment.  
Thus, our goal is to develop a method that can efficiently transfer a policy trained in a source environment to a target environment with minimal assumptions (i.e. no state and reward information), as illustrated in \myfigref{fig:teaser}.

However, RL policies trained in one environment tend to perform poorly in the other due to the \textit{visual domain gap} (e.g. lighting, viewpoint, clutter, and background) and \textit{dynamics domain gap} induced by physical differences (e.g. the mass of objects and calibration of the robot)~\citep{jakobi1995noise, tobin2017domain, peng2018sim}.
Hence, a variety of approaches propose to train a policy robust to domain gaps through noise injection~\citep{jakobi1995noise}, adversarial training~\citep{ganin2016domain, pinto2017robust, rajeswaran2017epopt}, learning domain-invariant representations~\citep{gupta2017learning}, or domain randomization~\citep{tobin2017domain, james2019sim, peng2018sim}. 
Yet, these approaches only transfer well within the training domain distribution, and training such robust policies becomes infeasible for large domain ranges. 
Instead, grounding-based policy transfer methods directly minimize the visual~\citep{rao2020rl, bousmalis2018using} and dynamics~\citep{hanna2017grounded, desai2020imitation} domain gap by grounding the source environment in the target environment.
Instead, domain adaptation methods learn explicit observation mappings~\citep{rao2020rl, bousmalis2018using} or action mappings~\citep{hanna2017grounded, desai2020imitation} to close the visual and dynamics domain gap respectively.  The mappings are then used to modify the domain of the policy or source environment to match that of the target environment.

These methods are designed to address only one type of domain gap at a time; however closing both visual and dynamics domain gaps is crucial for successful policy transfer as most robot learning scenarios include both. 

We propose \fullmethod~(\method) for policy transfer by modifying, or ``grounding'', the source environment to minimize both visual \textit{and} dynamics domain gaps to the target environment with minimal task information.
Concretely, we develop a novel policy transfer method via iterative environment grounding that alternates between (1) grounding the source environment in the target environment by learning both visual and dynamics transformations; and (2) training a policy on the grounded source environment, which has a smaller domain gap to the target environment while  still providing rich task supervision (e.g. reward and state information) from the source environment. 
Our iterative training gradually improves the grounded environment and the policy, leading to successful transfer to the target environment. For source environment grounding, we learn the visual and dynamics correspondences between the two environments from unpaired data from both domains via unsupervised correspondence learning algorithms. 
Note that our setup is robot learning friendly, as our method does not require any instrumentation for acquiring reward and state information in the target environment.

Our contributions are threefold: (1) we propose a grounded environment model that can handle both visual and dynamics domain gaps, (2) we propose \method, a novel policy transfer method with iterative environment grounding that gradually improves the grounded environment and policy, and (3) we develop a benchmark of locomotion and robotic manipulation tasks for policy transfer across both visual and dynamics domain gaps. On this benchmark, we demonstrate that our method can effectively transfer a policy to target environments with large visual and dynamics domain gaps where previous policy transfer approaches fail. 
\section{Related Work}

Efficient policy transfer between two different domains is a promising research direction for robot learning with applications in simulation-to-real and real-to-real transfer. However, the existence of visual and dynamics domain gaps between environments makes policy transfer challenging. The most naive approach is to finetune the policy in the target environment~\citep{rusu2017sim, julian2020finetune}. But, this is often not feasible if the reward function is not available at deployment and it requires a lot of expensive real-world interactions.

Instead of finetuning, \textbf{domain randomization} approaches randomize parameters of the source environment during training, resulting in a policy robust to a wide range of domains. 
Although domain randomization demonstrates promising results in manipulation~\citep{tobin2017domain, james2017transferring, peng2018sim} and locomotion~\citep{peng2020learning}, this approach is hardly applicable to real-to-real transfer scenarios, where the environment is not easily modifiable. It has also shown limited generalization as the domain gap becomes larger or if the target environment is outside of the training randomization range, as we show in \myfigref{fig:domain_exps}.
Similarly, a variety of approaches have been proposed to train a policy robust to domain gaps through adding random noise~\citep{jakobi1995noise}, learning domain-invariant features~\citep{gupta2017learning}, or adversarial training~\citep{ganin2016domain, pinto2017robust, rajeswaran2017epopt}. Yet, these approaches also work on a limited target domain distribution. 

Another avenue of work adapts a policy to a specific target domain by learning visual correspondences~\citep{bousmalis2018using, rao2020rl} to address the \textit{visual} gap across domains (e.g. simulation rendering to real-world image) .
Visual correspondences can be learned from unpaired data by optimizing cycle-consistency losses~\citep{zhu2017unpaired} with additional regularization, such as Q-function prediction from offline data~\citep{rao2020rl} or task success prediction~\citep{bousmalis2018using}.
However, even with the perfect visual correspondences, policy transfer can fail if the dynamics or physical properties of the target environment differ from those during training.

 To bridge the \textit{dynamics} gap between environments, prior approaches learn an action transformation to compensate the dynamics mismatch between domains based on learned dynamics models, explicitly~\citep{hanna2017grounded, desai2020stochastic, Malmir2020Robust} or implicitly~\citep{desai2020imitation, karnan2020reinforced}.
 One application of this action transformation is \textbf{environment grounding}, which grounds the source environment in the target environment, and trains a policy on the grounded environment.  The learned action transformations are used to modify the source environment such that the grounded source environment dynamics more closely match the target environment dynamics.  Then, a policy can be trained on this grounded environment, resulting in better transfer to the target environment. Yet, these methods only work when the source and target environments share the same state space, and hence cannot handle policy transfer with visual domain gaps. 

Recently, \citet{zhang2021learning} proposes to learn cross-domain correspondences for policy transfer across input modalities and physics differences, but their approach is limited to state-based policies in the source environment and is not suitable for transferring visual policies. In contrast, our method learns to close both the visual and dynamics domain gaps making policy transfer possible without the assumption of shared dynamics or the availability of a low-dimensional state representation.

\begin{figure*}[t]
    \centering
    \includegraphics[width=0.95\linewidth]{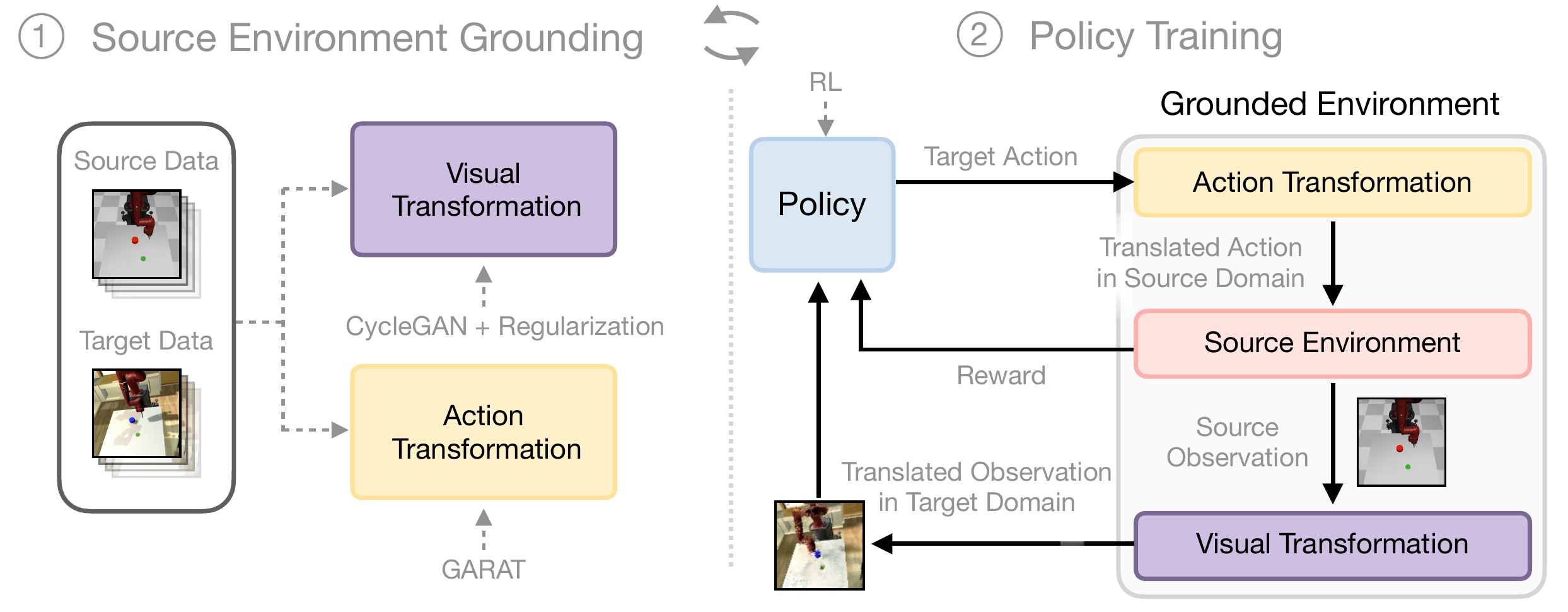}
    \caption{
        Our approach, IDAPT, alternates between two stages: (1)~source environment grounding and (2)~policy training. We ground the source environment in the target environment by learning visual (purple) and action (yellow) transformations. During the grounding stage, we first train the visual transformation on unpaired images, and then train the action transformation.  During policy training, we fix the grounded environment and optimize the policy using RL.
    }
    \label{fig:training}
\end{figure*}

\section{Approach}

In this paper, we aim to address the problem of transferring a policy from a well-instrumented and controlled environment to a target environment with minimal task information (i.e. no reward and state information). However, in many robotics setups, policies trained in the source environment struggle at performing the learned task in the target environment due to the visual and dynamics differences. To overcome such domain gaps, we introduce IDAPT, a novel policy transfer method with iterative environment grounding. The key goal is to learn a grounded source environment that mimics the target domains so that a policy trained in this grounded environment can then directly be executed in the target environment.

\subsection{Problem Formulation}
\label{sec:problem_formulation}

We consider an RL framework for policy transfer of a task from a source environment $S$ to target environment $T$.  The task is defined by a shared state space $\mathcal{S}$, action space $\mathcal{A}$, and reward function $\mathcal{R}: \mathcal{S} \times \mathcal{A} \rightarrow \mathbb{R}$.  The source and target environments have different visual observation spaces, $\mathcal{O}^S$ and $\mathcal{O}^T$, and transition functions, $\mathcal{T}^S: \mathcal{S} \times \mathcal{A} \rightarrow \mathcal{S}$ and $\mathcal{T}^T: \mathcal{S} \times \mathcal{A} \rightarrow \mathcal{S}$, respectively.  Note that the source environment provides full access to reward $r^S$, visual observation $o^S$, and state $s^S$, while the target environment only provides visual observations $o^T$.  Given a time horizon $H$ and discount factor $\gamma \in [0,1]$, our objective is to learn the optimal visual policy $\pi^*(a_t|o^T_t)$ that maximizes the expected return in the target environment:  
\begin{equation}
    \label{eqn:target_J}
    J(\pi) = \mathbb{E}_{\tau \sim p(\tau|\pi,\mathcal{T}^T )} \left [\sum_{t=0}^H \gamma^t r_t \right ].
\end{equation}
Due to the differences in observations and transition functions, this is not equivalent to maximizing the expected return in the source environment and results in different optimal policies.

In addition, we assume access to task-agnostic datasets of unpaired visual observations in both domains, $\mathcal{D}_o^S = \{(o_t^S, s_t^S), (o_{t+1}^S, s_{t+1}^S), \cdots \}$ and $\mathcal{D}_o^T = \{o_t^T, o_{t+1}^T, \cdots \}$. These task-agnostic datasets do not need to align with the current task, and therefore they can be collected via autonomous exploration, hand-specified calibration sequences, or policies trained for other tasks.  However, we assume the datasets share similar underlying state distributions, \ie they are collected by similar policies.

\subsection{Iterative Domain Alignment for Policy Transfer}
\label{sec:iterative_domain_alignment}
To overcome domain gaps for policy transfer, IDAPT iterates between (1) source environment grounding; and (2) policy training, as illustrated in \myfigref{fig:training}:
\begin{itemize}
    \item \textbf{Source environment grounding}: we learn a \textit{visual transformation} and \textit{action transformation} that compensate for the visual and dynamics domain gaps between the two environments from unpaired trajectories.
    \item \textbf{Policy training}: the grounded environment acts as a proxy for the target environment, providing interactions with reward to train a policy.
\end{itemize}

We first initialize the grounded environment by training the visual transformation on the unpaired, task-agnostic dataset of images and initializing the action transformation as the identity function. From the following grounding step, we collect task-relevant trajectories using the learned policy to improve both transformations. For every environment grounding iteration, we further train our policy in the improved grounded environment.
Since we do not have access to paired data or expert task data, the transformations for grounding are trained with the unpaired, sub-optimal data. However, as the learned policy generates still sub-optimal but more task-relevant data, the visual and dynamics transformations become more accurate around the task-relevant state and action spaces. The improved grounding improves policy transfer as the domain gaps between the grounded environment and the target environment shrink. Therefore, we iterate over these two stages to gradually improve the alignment between the grounded and target environments and the performance of the policy. The entire training procedure is outlined in \myalgref{alg:method}.

\subsection{Learning Grounded Environment}
\label{sec:grounded_environment}

The goal of this step is to ground the source environment \textit{both visually and physically} using unpaired source and target environment trajectories. 
As the grounded environment is closer to the target environment, the policy trained in the grounded environment transfers better than one trained in the original source environment.

As illustrated in \myfigref{fig:training} (right), our grounded source environment is composed of three components: the source environment $S$, visual transformation $G: \mathcal{O}^S \rightarrow \mathcal{O}^T$, and action transformation $F: \mathcal{O}^T \times \mathcal{A} \rightarrow \mathcal{A}$. 
With these transformations, we ground the source environment by:
\begin{enumerate}[label=(\arabic*)]
\item Translating the source observation to the target domain, $\hat{o}_t^T = G(o_t^S)$.
\item Translating the target action to the source action to compensate dynamics mismatches, $\hat{a}_t^S = F(\hat{o}_t^T, a_t^T)$.
\item Rolling out the source environment, $o_{t+1}^S = \mathcal{T}^S(o_t^S, \hat{a}_t^S)$. 
\item Translating the next source observation to the target domain, $\hat{o}_{t+1}^T = G(o_{t+1}^S)$. 
\end{enumerate}
Through this process, the grounded environment can simulate the target environment by taking $a_t^T$ and providing $\hat{o}_{t+1}^T$. In addition, this grounded environment can provide the task reward $\hat{r}_t^T = r_t^S$, which is not available in the target environment.

To achieve a transferable policy, it is crucial for the grounded environment to simulate the target environment as closely as possible. In the rest of this section, we explain how to efficiently learn accurate grounding by learning a visual transformation and action transformation from limited target environment data.  For each grounding step, we train the transformations with 1k target environment samples.

\subsubsection{\textbf{Learning Visual Transformation}} 
\label{sec:visual_transformation}

To transfer a visual policy, the visual domains during training and deployment should be similar. To match the domains, we ground the source environment to be visually similar to the target environment by learning a source-to-target visual transformation $G$. 

The visual transformation is first initialized using the task-agnostic datasets $\{\mathcal{D}_o^S, \mathcal{D}_o^T\}$, and then we iteratively improve it with online datasets $\{\mathcal{D}^S_{online}, \mathcal{D}^T_{online}\}$ collected using the current policy.  This serves to improve the transformation in the task-relevant observation spaces which may not be well represented in the initial task-agnostic datasets.

We train our visual transformation using an unsupervised image-to-image translation method, CycleGAN~\citep{zhu2017unpaired}, which optimizes the cycle-consistency loss between two domains. However, due to the lack of paired images, the resulting visual transformation can map semantically incorrect images, \eg change the arrangements of objects in the scene.
To ensure the state information is preserved across domains (i.e. the domains are semantically aligned), we propose a \textit{state reconstruction regularization}, which encourages the source state and the predicted state from the translated observation to be the same. Note that the state reconstruction can be replaced with any form of self-supervision, such as a robot state or reward.

\begin{figure}[t]
    \centering
    \includegraphics[width=\linewidth]{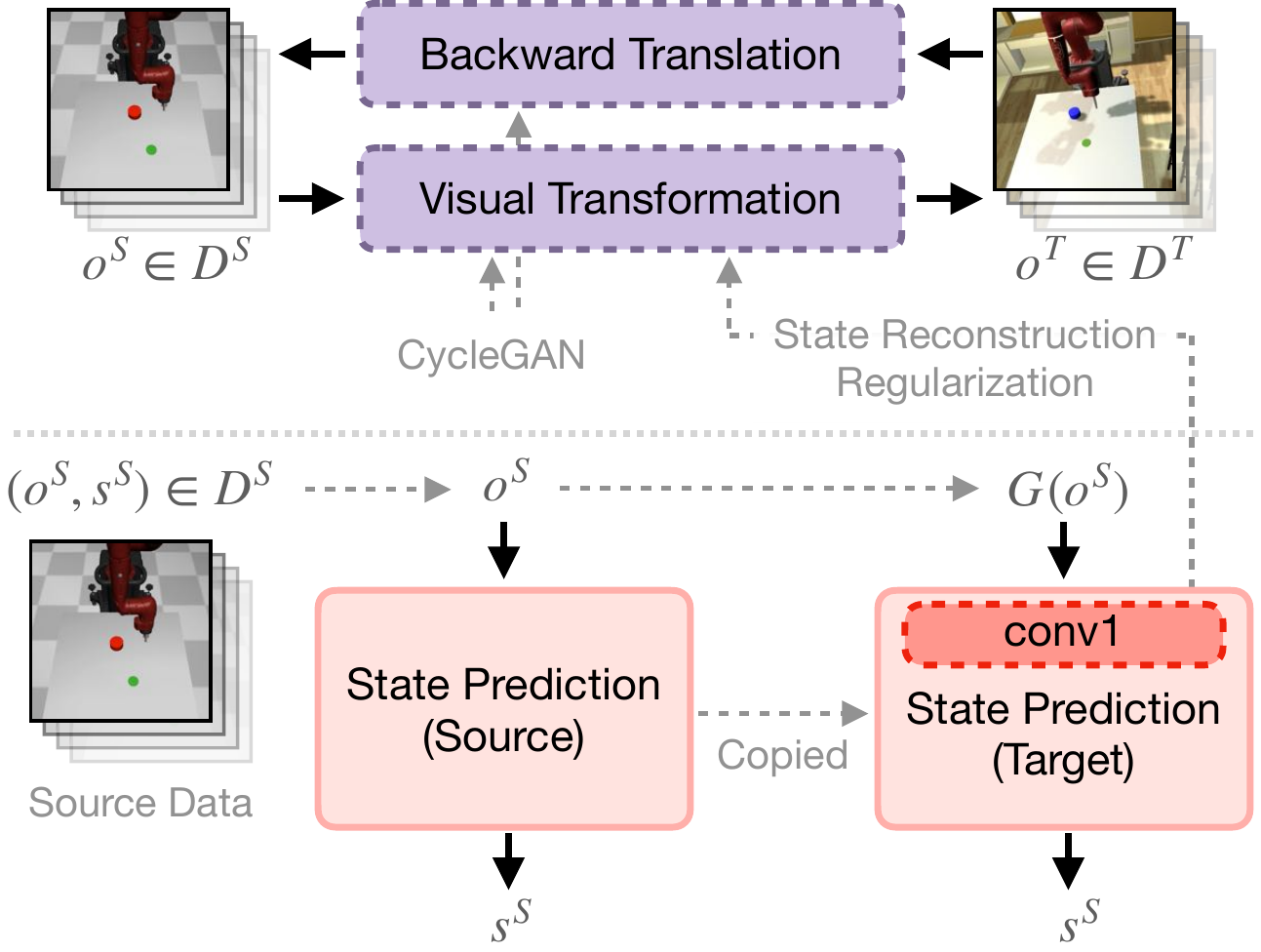}
    \caption{
        To train the visual transformation, we first train the source state predictor (left bottom) with image-state pairs from the source environment, and copy its weights to the target state predictor. Then, we train image translation models (purple) and the first layer (\texttt{conv1}) of the target state predictor jointly using the CycleGAN loss and state reconstruction regularization.
    }
    \label{fig:visual_transformation}
\end{figure}

Since the target environment does not provide the state information, we cannot directly train a target state predictor. Instead, we first train a source state predictor $s = P^{S}(o^{S})$ using the source domain dataset $\mathcal{D}_o^S$ with state labels, and generate pseudo-labeled target domain dataset $\{(G(o^S), s)|(o^S, s) \in \mathcal{D}_o^S \}$ to train the target state predictor $s = P^{T}(o^{T})$. Using this target state predictor, we can compute the state reconstruction regularization loss $\lVert P^{T}(G(o^{S})) - s \rVert_1$. We can jointly train the visual transformation and target state predictor by optimizing the CycleGAN objective with state reconstruction regularization:
\begin{equation}
\begin{aligned}
    \label{eqn:visual_loss}
    \mathcal{L}_{visual} &= \mathcal{L}_{CycleGAN}(o^{S}, o^{T}) + \lambda \lVert P^{T}(G(o^{S})) - s \rVert_1,
\end{aligned}
\end{equation}
where $\lambda$ is a weighting factor for the regularization. For correct visual alignment, we encourage the state predictors to extract the shared state information: we initialize the target state predictor with the weights of the source state predictor and finetune only the top layer (\texttt{conv1}) in the target domain~\citep{aytar2017crossmodal, jeong2020selfsupervised}, as illustrated in \myfigref{fig:visual_transformation}.  During subsequent iterations, we follow the same procedure of first training $P^{S}$ and copying its weights to $P^{T}$, then jointly training the rest of the model to finetune to task-relevant data.

\subsubsection{\textbf{Learning Action Transformation}}
\label{sec:action_transformation}

In addition to the visual domain gap, physical differences (e.g. friction, mass, robot calibration) hinder policy transfer to the target domain.  To deal with the discrepancy between the dynamics of the source and target environments, we learn an action transformation~\citep{hanna2017grounded, desai2020stochastic, karnan2020reinforced, desai2020imitation} that compensates for the dynamics mismatch.
We learn the action transformation $F(a^{S}|o^T,a^{T})$ from a target domain observation-action pair to a source domain action such that the resulting transition in the respective environments is the same.  This grounds the source environment in the target environment's dynamics, and thus a policy trained in the grounded environment can generate the same trajectories in the target environment.  

However, prior grounded action transformation approaches require access to a shared, low-dimensional state space across domains and are not suited for visual observations with domain gaps. Since our goal is to solve more realistic cases where the target environment does not provide access to such information, we extend the grounded action transformation to image observations with visual domain differences.  We use the visual features $f_\pi(o^T)$ generated by the policy network (i.e. output of the convolutional layers of the policy network) as a proxy for the state since the policy extracts low-dimensional representations that contain task-relevant information.  Since $f_\pi$ is trained on the target domain we can directly use it to encode target domain observations into proxy states.  In the source domain, we can use the visual transformation $G$ to obtain the corresponding target domain observation, $f_\pi(G(o^S))$. Hence, we can use $f_\pi$ to generate a shared feature space to act like proxy states for both source and target domain observations.

To efficiently learn an action transformation on our proxy state representation, we use the state-of-the-art grounded action transformation method, GARAT~\citep{desai2020imitation} -- which requires only a few target domain trajectories -- by framing our problem as an adversarial imitation learning from observation problem in the source environment.  Specifically, the action transformation, recast as the agent, transforms actions between domains such that the resulting transitions in the source environment resemble the transitions gathered in the target environment, thereby correcting for any dynamics differences.   
We implement GARAT using GAIfO~\citep{torabi2018generative} and PPO~\citep{PPO}.

\begin{algorithm}[t]
    \caption{\fullmethod}
    \label{alg:method}
    \begin{algorithmic}[1]
        \STATE \textbf{Input:} Task-agnostic dataset $\{\mathcal{D}_o^S,\mathcal{D}_o^T\}$, number of grounding steps $N$
        \STATE Randomly initialize policy $\pi$ and visual transformation $G$
        \STATE Initialize action transformation $F$ as identity function
        \STATE Pretrain source state predictor $P^S$ on $\mathcal{D}^S_o$
        \STATE Optimize $G$ with $P^T$ to minimize \myeqref{eqn:visual_loss} on $\{ \mathcal{D}^S_o, \mathcal{D}^T_o \}$
        \FOR {$i = 1,2,\cdots,N$}
            \STATE Optimize $\pi$ in grounded environment with RL 
            \STATE Roll out $\pi$ in target environment to obtain $ \mathcal{D}^T_{online}$
            \STATE Roll out $\pi$ in grounded environment to obtain $ \mathcal{D}^S_{online}$
            \STATE Finetune $P^S$, $G$, and $P^T$ on $\{\mathcal{D}^S_{online},\mathcal{D}^T_{online}\}$
            \STATE Optimize $F$ with GARAT on $\mathcal{D}^T_{online}$
        \ENDFOR
        \STATE \textbf{Output:} Policy $\pi$ to deploy in target environment
    \end{algorithmic}
\end{algorithm}

\subsection{Policy Training}
\label{sec:policy_training}

With the grounded environment, a policy $\pi(a^T | o^T)$ can be trained using any RL algorithm as if it were trained on the target environment. The policy learns to maximize the expected return  from the grounded environment:
\begin{equation}
    \label{eqn:grounded_J}
    \hat{J}(\pi) = \mathbb{E}_{\tau \sim p(\tau| F(G, \pi \circ G),\mathcal{T}^S )} \left [\sum_{t=0}^H \gamma^t \hat{r}_t \right ].
\end{equation}
In this work, we use Asymmetric SAC~\citep{pinto2017asymmetric} for policy training, a variant of SAC~\citep{haarnoja2018sac} that is efficient for learning an image-based policies by using state-conditioned critics.  

Even though the policy only learns from the grounded source environment interactions, we can directly execute this policy on the target environment as our learned transformations effectively close the domain gaps between the source and target environments. This makes IDAPT well suited for cases where task supervision is not available and data collection is expensive or dangerous in the target environment. Instead, IDAPT efficiently learns the visual transformation and action transformation using a few target domain interactions, and then trains a policy by fully utilizing rich, cheaply obtained data from the well instrumented and controlled source environment.

The initial task-agnostic dataset may not be sufficient to train a grounded environment accurate enough to learn a transferable policy. 
Hence, we improve the grounded environment using the task-relevant data collected by the updated policy, and then train the policy again using the better grounded environment. We iterate between these two stages to gradually improve the grounded environment and the policy. The entire training procedure is outlined in \myalgref{alg:method}.

\section{Experiments}

In this paper, we propose a policy transfer method with iterative grounding across visual and dynamics domains without task supervision in the target environment. IDAPT iteratively learns visual and action transformations with limited target environment data, uses these transformations to ground the source environment such that it mimics the target environment, and then trains a transferable policy in the grounded environment with rewards from the source environment.

Through our experiments, we aim to answer the following questions: (1)~Can IDAPT effectively transfer a policy across visual and dynamics differences with minimal target environment interactions? (2)~How does IDAPT scale to wider domain gaps when compared to prior work?  (3)~Can the iterative training process overcome poor initial datasets?

\subsection{Experimental Setup}

\begin{figure}[t]
    \centering
    \begin{subfigure}[t]{\linewidth}
        \centering
        \makebox[0.27\linewidth]{Source}
        \makebox[0.27\linewidth]{Target-easy}
        \makebox[0.27\linewidth]{Target-hard}
    \end{subfigure}
    \\
    \vspace{0.3em}
    \begin{subfigure}[t]{\linewidth}
        \centering
        \includegraphics[width=0.27\linewidth]{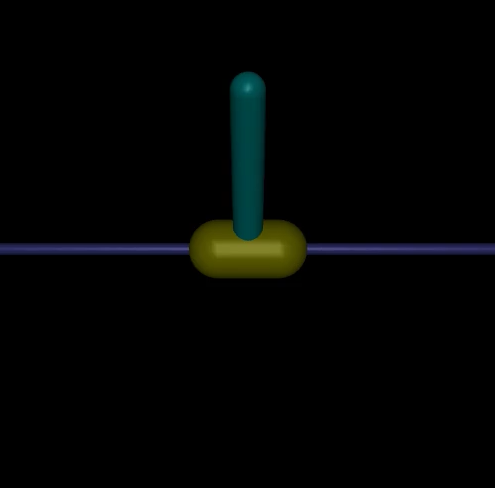}
        \includegraphics[width=0.27\linewidth]{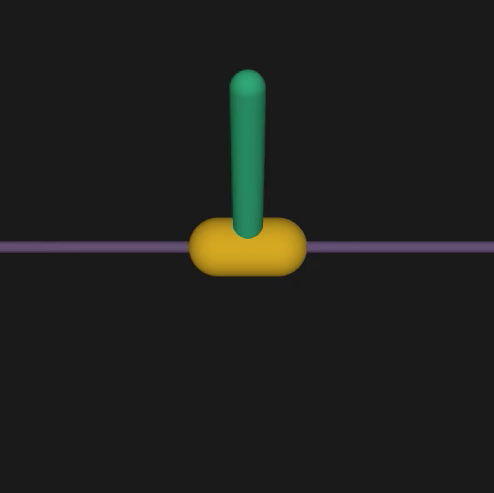}
        \includegraphics[width=0.27\linewidth]{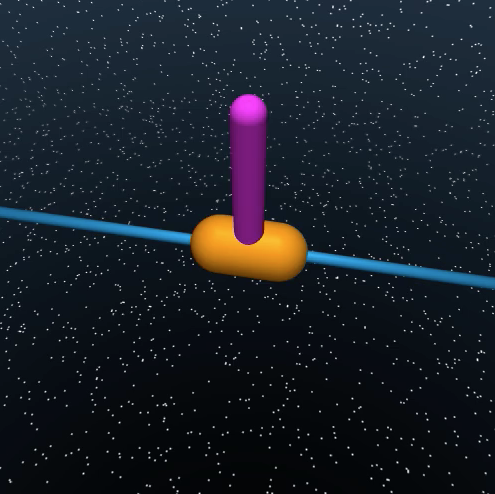}
        \caption{InvertedPendulum}
    \end{subfigure}
    \\
    \begin{subfigure}[t]{\linewidth}
        \centering
        \includegraphics[width=0.27\linewidth]{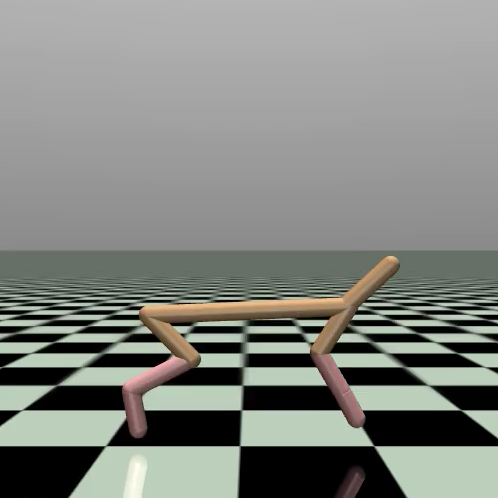}
        \includegraphics[width=0.27\linewidth]{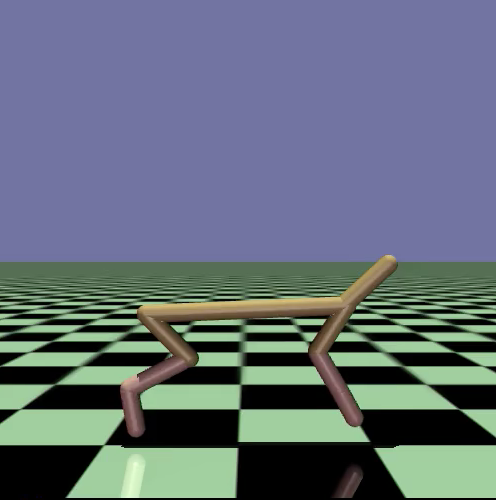}
        \includegraphics[width=0.27\linewidth]{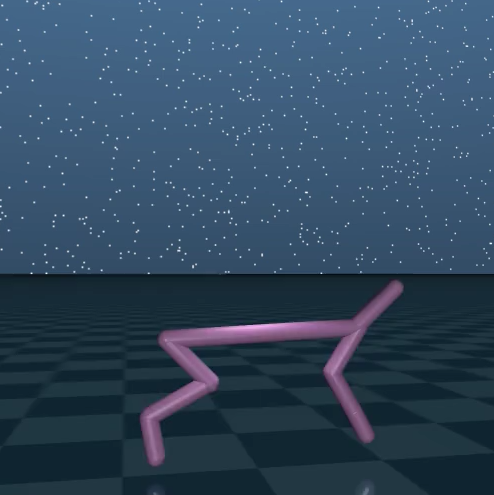}
        \caption{HalfCheetah}
    \end{subfigure}
    \\
    \begin{subfigure}[t]{\linewidth}
        \centering
        \includegraphics[width=0.27\linewidth]{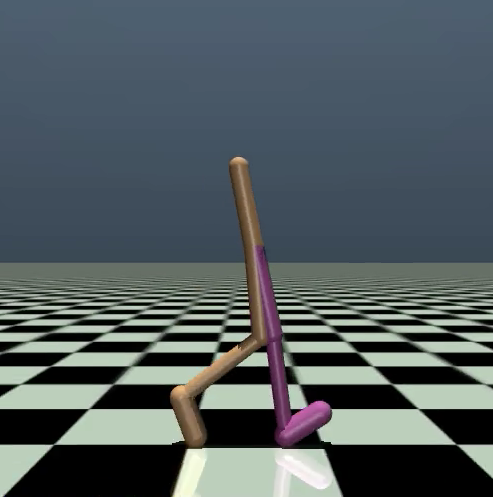}
        \includegraphics[width=0.27\linewidth]{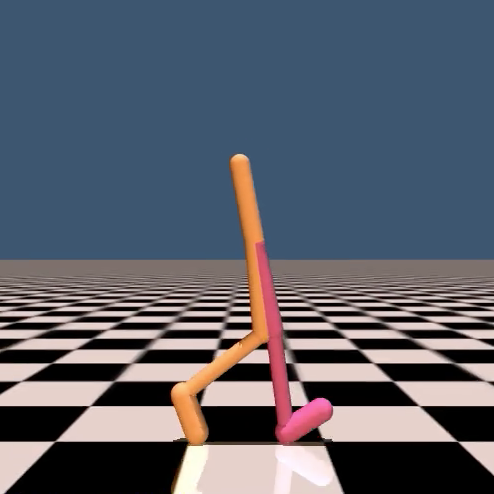}
        \includegraphics[width=0.27\linewidth]{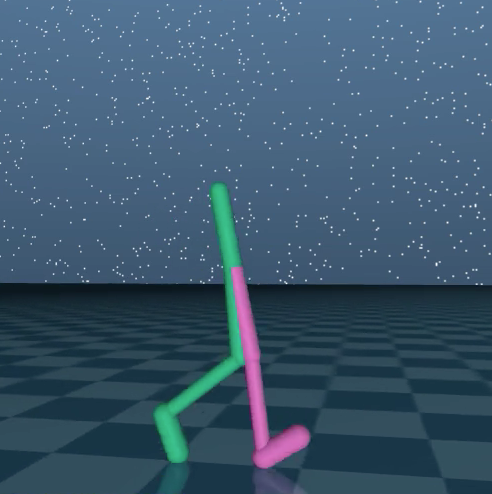}
        \caption{Walker2d}
    \end{subfigure}
    \\
    \begin{subfigure}[t]{\linewidth}
        \centering
        \includegraphics[width=0.27\linewidth]{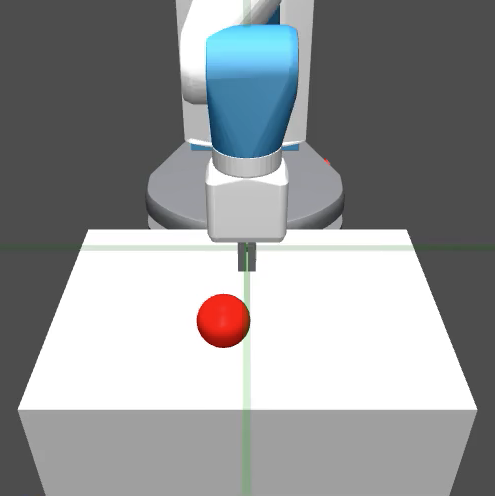}        
        \includegraphics[width=0.27\linewidth]{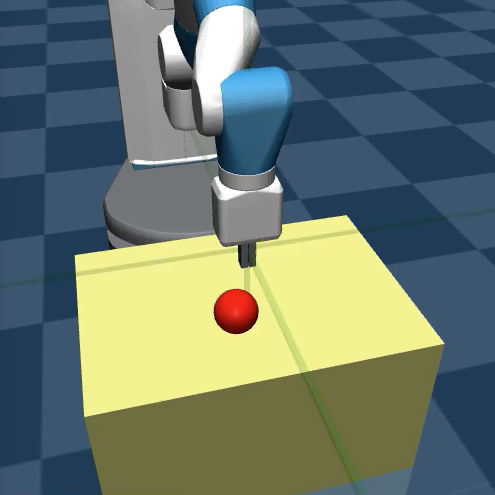}        
        \includegraphics[width=0.27\linewidth]{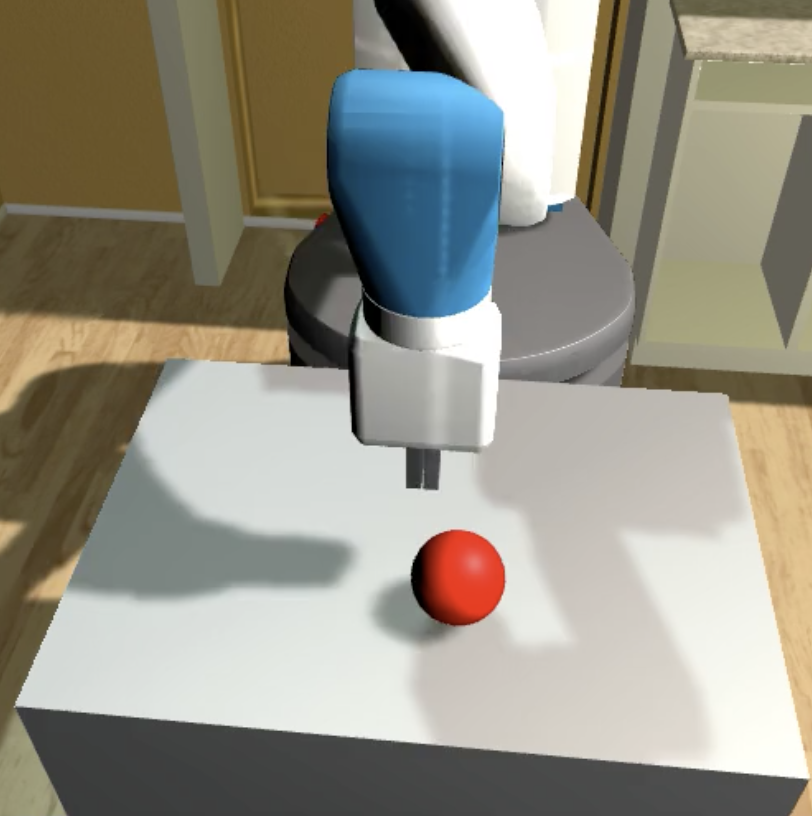}
        \caption{Fetch-Reach}
    \end{subfigure}
    \\
    \begin{subfigure}[t]{\linewidth}
        \centering
        \includegraphics[width=0.27\linewidth]{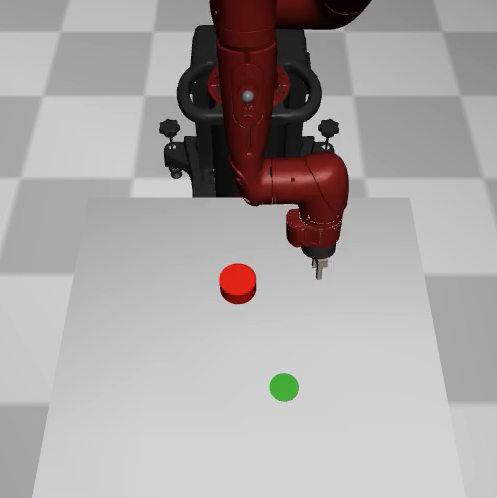}        
        \includegraphics[width=0.27\linewidth]{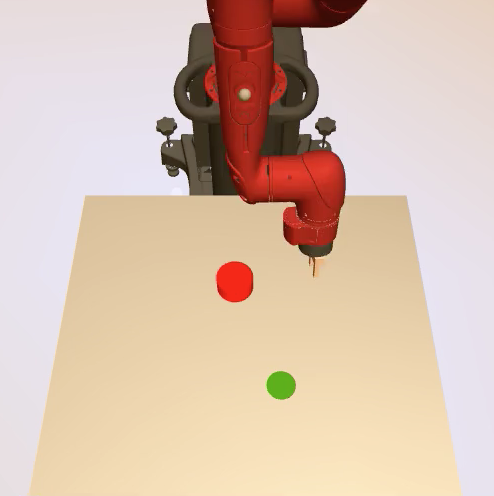}        
        \includegraphics[width=0.27\linewidth]{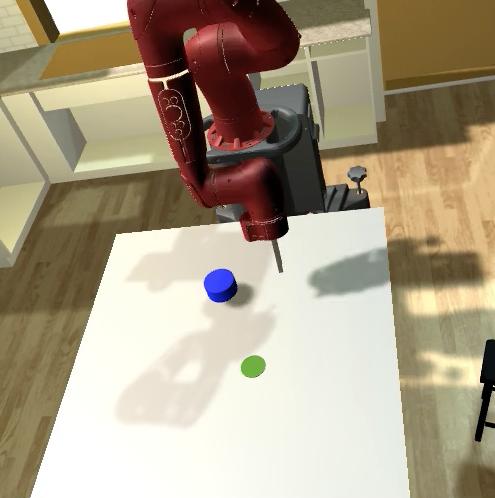}
        \caption{Sawyer-Push}
    \end{subfigure}
    \caption{
        Visualize source domain (left) and target domains with small (middle) and large (right) domain gaps. While the target-easy environments slightly differ in color and lighting condition, the target-hard environments include drastic changes in camera viewpoint, texture, and background. Especially in the manipulation tasks, we use the Unity3D rendering engine to make the target-hard environments look more realistic. 
    }
    \label{fig:tasks}
\end{figure}

To test how well IDAPT transfers a policy with various visual and physical domain gaps, we design two benchmark target environments, one with smaller domain gaps (Target-easy) and one with larger domain gaps (Target-hard), on five simulation-to-simulation transfer tasks: classic inverted pendulum, two locomotion, and two robotic manipulation tasks. We vary both physical parameters and visual appearances for the source and target environments.  The changes are summarized in \mytbref{tab:dynamics-diff} and \myfigref{fig:tasks}.  

For the evaluation metric of a policy transfer, we use the target domain performance (reward or success rate).  For all methods and tasks, we report the average target environment performance and standard deviation over 5 random seeds unless otherwise stated.


\subsection{Baselines}

To provide baselines for policy transfer performance, we evaluate multiple representative policy transfer methods that can handle visual and dynamics domain gaps as following:
\begin{itemize}
    \item \textbf{Domain Randomization (DR)}~\citep{tobin2017domain, peng2018sim} learns robust policies by randomizing the source environment's visual and dynamics parameters, including color, texture, lighting, viewpoint, friction, and armature. In order to provide a fair comparison, we define the randomization ranges to include both source and target environments (\textbf{DR-Wide}). For the cases where the DR policy is unable to learn to cover the full range, we also include a policy trained with a smaller randomization range (\textbf{DR-Narrow}) that only includes source and target-easy domains.
    
    \item \textbf{Adaptive RL} learns a robust adaptive policy under domain randomized environments that can identify and adapt to the current domain on the fly.  This allows the policy to be more flexible and accommodate a wider training range of domains.  We implement this with an LSTM-based policy.

    \item \textbf{Learning Cross-Domain Correspondence (CC)}~\citep{zhang2021learning} utilizes a learned observation mapping and an action mapping to transfer a policy across input modalities and dynamics domains.  CC addresses difference in input modalities instead of visual domain gap so we use two different modes for an informative comparison.  \textbf{CC-State} learns a state-to-state observation mapping to transfer a state-based policy across dynamics differences.  \textbf{CC-Image} learns a state-to-image observation mapping to transfer a policy across input modalities and dynamics domains.  CC uses a dataset of 50k samples from both domains.  We additionally make CC iterative by gathering 1k target environment samples using the trained mappings between every iteration.
    
    \item \textbf{\method} (our method) takes iterative grounding and policy training steps, as described in \mysecref{sec:iterative_domain_alignment}. For these experiments, we start out with a dataset of 20k images from both domains and additionally collect 1k target environment samples per grounding step.  We execute 5 grounding steps for target-hard environments and 1 grounding step for target-easy environments.  This results in a total data usage of 25k interactions which is 50\% of that of the CC baselines.  
\end{itemize}

For further implementation details, please refer to appendix, \mysecref{sec:environment_details} for environments, \mysecref{sec:baseline_implementation} for baseline implementations, \mysecref{sec:idapt_details} for our method.

\begin{table}[t]
\centering
\begin{tabular}{ ccccc } 
 \toprule
 \multirow{2}{7.5em}{\centering Task} & \multirow{2}{6.5em}{\centering Parameter} & \multirow{2}{3em}{\centering Source} & \multicolumn{2}{c}{Target} \\ 
 & & & Easy & Hard \\
 \midrule
 InvertedPendulum & Pendulum mass & 4.895 & 50 & 200 \\ 
 \midrule
 HalfCheetah & Armature & 0.1 & 0.18 & 0.4 \\ 
 \midrule
 Walker2d & Torso mass & 3.534 & 5.4 & 10.0 \\ 
 \midrule
 \multirow{2}{7.5em}{\centering Fetch-Reach} & Action Rotation & 0 & 30 & 45$^{\circ}$\\
 & Action Bias & 0 & 0 & -0.5 \\
 \midrule
 Sawyer-Push & Puck mass & 0.01 & 0.03 & 0.05 \\
 \bottomrule
\end{tabular}
\vspace{1em}
\caption{Dynamics differences between source and the two target domains.  We chose different physical parameters to vary for each environment that would affect transfer performance.}
\label{tab:dynamics-diff}
\end{table}

\begin{figure*}[t]
    \centering
    \begin{subfigure}[t]{0.29\linewidth}
        \centering
        \includegraphics[width=\linewidth]{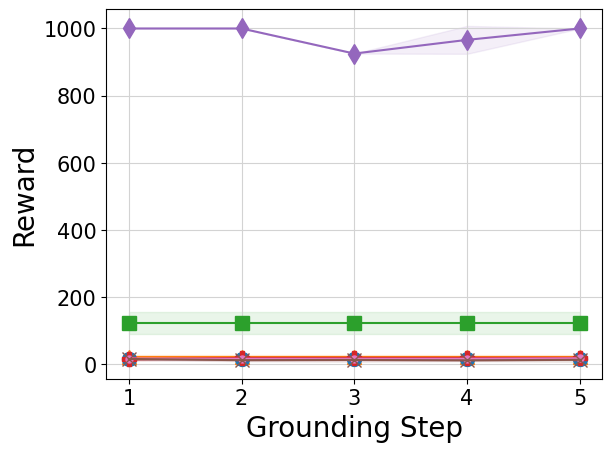}
        \caption{InvertedPendulum}
        \label{fig:result_step:invertedpendulum}
    \end{subfigure}
    \begin{subfigure}[t]{0.29\linewidth}
        \centering
        \includegraphics[width=\linewidth]{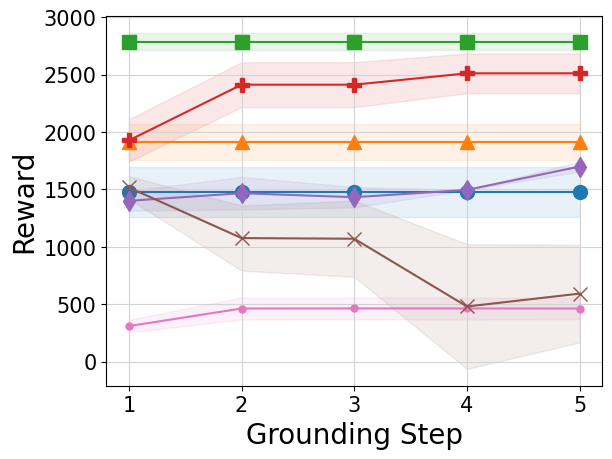}
        \caption{HalfCheetah}
        \label{fig:result_step:halfcheetah}
    \end{subfigure}
    \begin{subfigure}[t]{0.29\linewidth}
        \centering
        \includegraphics[width=\linewidth]{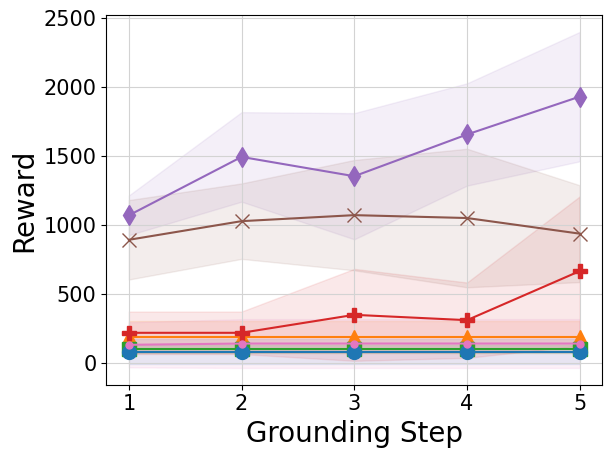}
        \caption{Walker2d}
        \label{fig:result_step:walker2d}
    \end{subfigure}
    \\
    \begin{subfigure}[t]{0.29\linewidth}
        \centering
        \includegraphics[width=\linewidth]{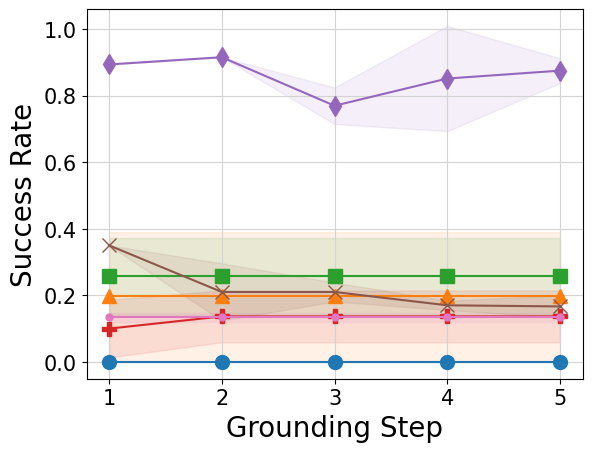}
        \caption{Fetch-Reach}
        \label{fig:result_step:fetchreach}
    \end{subfigure}
    \begin{subfigure}[t]{0.29\linewidth}
        \centering
        \includegraphics[width=\linewidth]{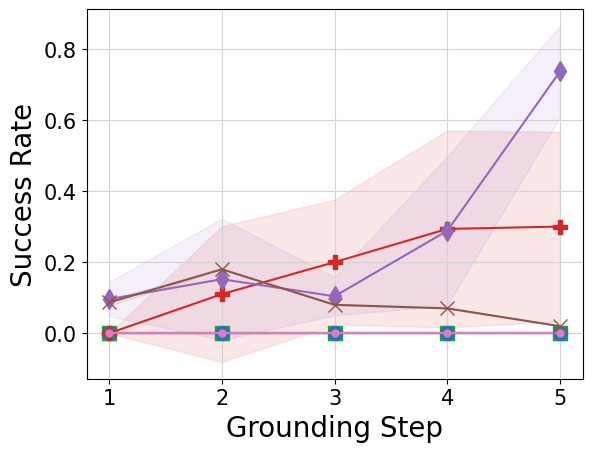}
        \caption{Sawyer-Push}
        \label{fig:result_step:sawyer_push}
    \end{subfigure}
    \begin{subfigure}[t]{0.29\linewidth}
        \centering
        \includegraphics[width=0.7\linewidth]{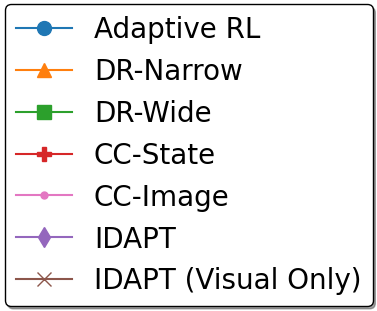}
    \end{subfigure}
    \caption{
    Performance of policies when evaluated in the target-hard domain.  We report success rates for manipulation tasks and reward for other tasks. For CC and \method, we report performance after each training iteration.  The remaining methods do not take online target environment interactions so we report their final evaluation performance after training.  For IDAPT, we use a backwards walking dataset for (b), (c) and randomly collected datasets for all other tasks. IDAPT (Visual Only) ablates our action transformation (over 3 random seeds). We evaluate the mean performance over 100 episodes for all methods. The results with normalized rewards can be found in appendix, \myfigref{fig:result_step_norm}.
    }
    \label{fig:result_step}
\end{figure*}

\subsection{Inverted Pendulum}

We first examine the InvertedPendulum task, which is a classic continuous control task~\citep{brockman2016openai}. We create the target environments by changing the pendulum mass,  color, and background color. For the target-hard environment, we make these changes more significant, tilt the camera angle, and change the background texture. We collect trajectories of 20k images by taking random actions for the task-agnostic dataset.

\begin{figure*}[t]
    \centering
    \begin{subfigure}[t]{0.3\linewidth}
        \includegraphics[width=\linewidth]{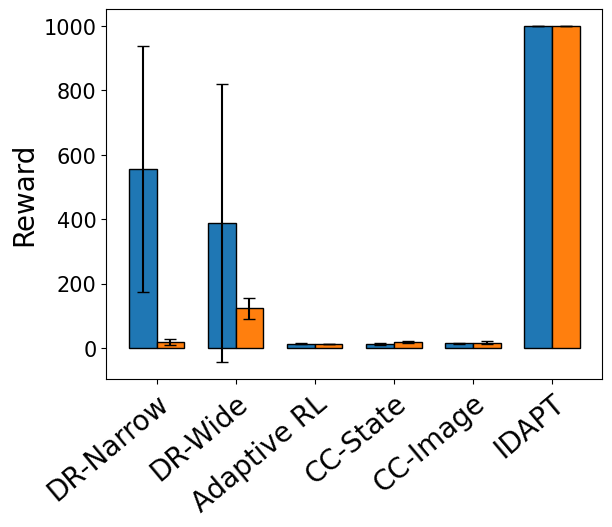}
        \caption{InvertedPendulum}
        \label{fig:result_all:invertedpendulum}
    \end{subfigure}
    \begin{subfigure}[t]{0.3\linewidth}
        \includegraphics[width=\linewidth]{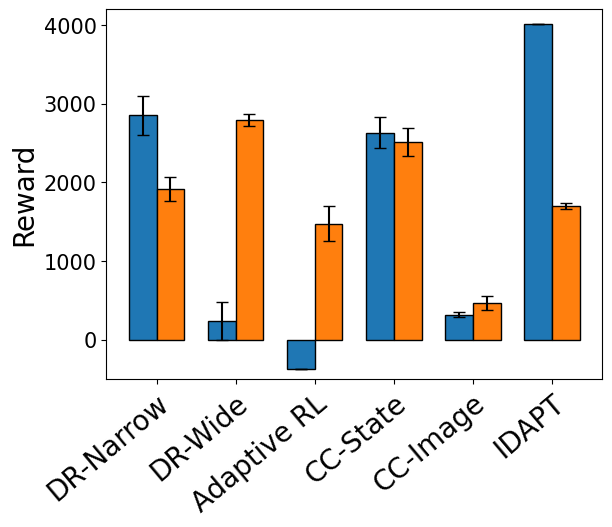}
        \caption{HalfCheetah}
        \label{fig:result_all:halfcheetah}
    \end{subfigure}
    \begin{subfigure}[t]{0.3\linewidth}
        \includegraphics[width=\linewidth]{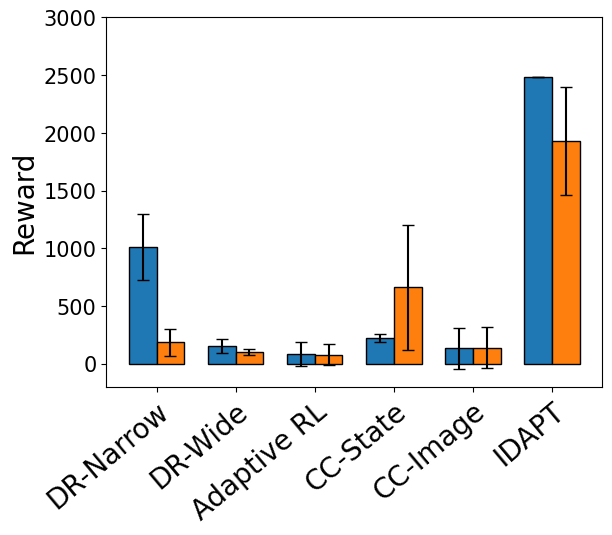}
        \caption{Walker2D}
        \label{fig:result_all:walker2d}
    \end{subfigure}
    \\
    \begin{subfigure}[t]{0.3\linewidth}
        \includegraphics[width=\linewidth]{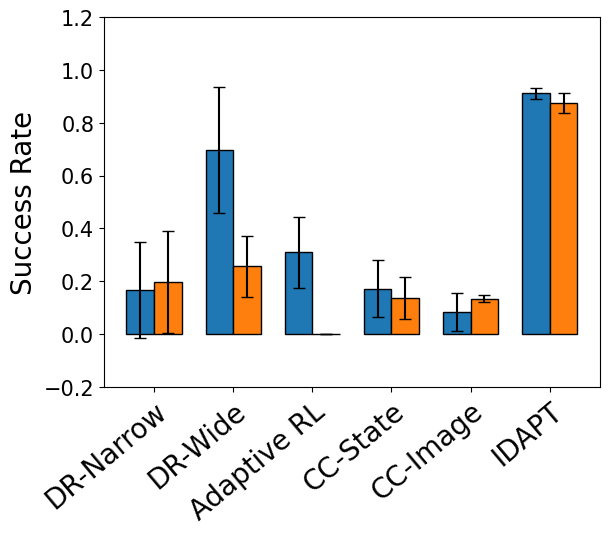}
        \caption{Fetch-Reach}
        \label{fig:result_all:fetch_reach}
    \end{subfigure}
    \begin{subfigure}[t]{0.3\linewidth}
        \includegraphics[width=\linewidth]{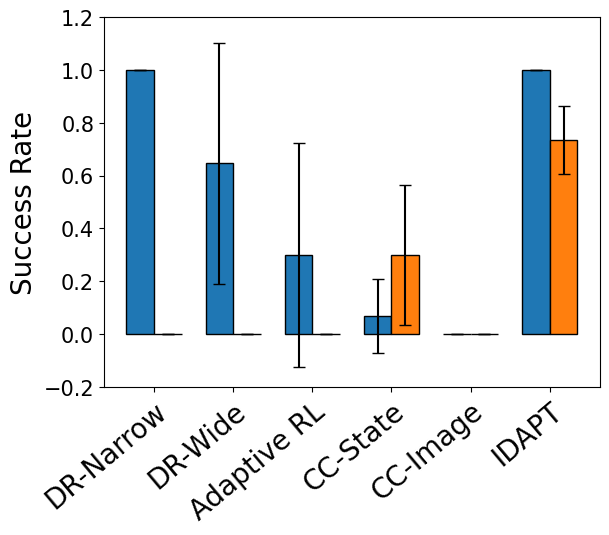}
        \caption{Sawyer-Push}
        \label{fig:result_all:sawyer_push}
    \end{subfigure}
    \begin{subfigure}[t]{0.29\linewidth}
        \centering
        \hspace{-5mm}
        \raisebox{95pt}{\includegraphics[width=0.6\linewidth]{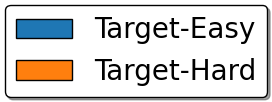}}
    \end{subfigure}
    \caption{
        Performance of policies transferred to the target domains with different domain gap sizes, target-easy and target-hard. Our proposed approach, IDAPT, outperforms the baselines on most environments and scales better to larger domain gaps.
    }
\label{fig:domain_exps}
\end{figure*}

\myfigref{fig:result_all:invertedpendulum} shows that our method outperforms baseline methods in transferring to both target-easy and target-hard domains. Moreover, IDAPT shows a high target data efficiency as it achieves nearly maximum reward starting from the first grounding step, as shown in \myfigref{fig:result_step:invertedpendulum}. 

We can also observe that DR-Narrow and DR-Wide achieve 60\% and 40\% success rates on the target-easy domain but fail to generalize to the target-hard domain.  Our results show that with the smaller randomization range, DR-Narrow learns the task well but fails to generalize beyond the training domain distribution. In contrast, DR-Wide is difficult to train as the resulting policy must learn to cover a large range of visual and physical domains; but once trained, it can perform better in the target environment with a large domain gap.

\subsection{Locomotion}

HalfCheetah and Walker2d are two representative locomotion tasks~\citep{brockman2016openai}. For dynamics domain shifts, we vary the armature~\citep{zhang2021learning} and torso mass~\citep{desai2020imitation} for HalfCheetah and Walker2d, respectively. For the visual domain shifts of the target-easy environment, we make small changes in the agent and background color. For the target-hard environment, we create a domain gap in the visual style of DeepMind Control Suite~\citep{tassa2018deepmind} and change the camera viewpoint.

To study the effect of different datasets for initial visual transformation training for IDAPT and correspondence training for CC, we collect two task-agnostic datasets, one using random actions and one with a policy trained to walk backwards. In the following results, we first compare absolute transfer performance by reporting the best performance out of the two datasets (random for CC, backwards for ours).  We then do an analysis on the effect of dataset quality in \mysecref{sec:dataset_analysis}.

In the target-hard environment of HalfCheetah, DR-Wide performs better than IDAPT (see \myfigref{fig:result_step:halfcheetah}), while Adaptive RL performs comparatively.  We hypothesize that the stable agent pose and relative ease of the task can allow a policy to find conservative actions that work with a large range of dynamics. On the other hand, learning explicit domain correspondences can still have some errors, resulting in poorer policy performance.  While CC-State also outperforms IDAPT, the performance difference to CC-Image demonstrates the added difficulty of the visual domain gap.

The Walker2D character is an example of a locomotion problem that is less stable and more prone to falling.  Here, we demonstrate the relative advantage of IDAPT for more difficult control tasks where domain differences are more crippling.  IDAPT is able to achieve up to 75\% of the upper bound value \myfigref{fig:result_step_norm:walker2d} while the baselines cannot improve much beyond direct transfer.  In fact, the strongest baseline here is DR-Narrow, but it does not generalize well to the target-hard task outside of its training distribution.  In contrast, DR-Wide fails to learn a good policy for any domain because the randomization range required to fit the target environment is too challenging for this type of control task.

\subsection{Manipulation}

We evaluate our method in two robotic manipulation tasks with the 7-DoF Fetch robot: Fetch-Reach~\citep{plappert2018robotics} and the 7-DoF Rethink Sawyer robot: Sawyer-Push. The robot must move its end effector to a goal position in Fetch-Reach or push a puck to a target position for Sawyer-Push. For the target-easy domain, we simply change colors, lighting conditions, and viewpoint (Fetch-Reach only); but in the target-hard domain, we emulate a realistic visual gap by generating more realistic backgrounds using Unity3D and changing viewpoint.  The dynamics differences for Sawyer-Push come from puck mass. In Fetch-Reach we bias and rotate the input actions to model calibration error.  For the initial target domain dataset, we collect 20k images from either domain with random end-effector control.

In Fetch-Reach, IDAPT achieves greater than 90\% success rate in both domain gaps (see \myfigref{fig:result_all:fetch_reach}), demonstrating a robustness to a wide visual gaps.  The visual transformation can handle drastic changes in viewpoint and background, as well as the more realistic rendering.  In addition, we attain a good success rate in just one grounding step (see \myfigref{fig:result_step:fetchreach}), showing the effectiveness of our grounded environment with only a few target environment interactions.

In Sawyer-Push, IDAPT achieves a 75\% success rate in five grounding steps. This task is more difficult due to the wide visual domain gap as well as the varying mass of the object which can be hard to adapt to based on task-agnostic data.  The increasing learning curve in \myfigref{fig:result_step:sawyer_push} shows the benefit of iterative domain alignment: IDAPT shows low performance for the first three grounding steps but as the visual and action transformations provide better grounding of the source environment, the policy successfully transfers to the target environment.  Moreover, in the initial dataset, the robot is very unlikely to have moved the puck, making it difficult for CC and IDAPT to model the puck dynamics or generate the correct visual translation, therefore failing to transfer a policy.  Both iterative CC and IDAPT are improved over multiple iterations as the transformation is finetuned over a better data distribution.

\begin{figure}[t]
    \centering
    \begin{subfigure}[t]{0.6\linewidth}        
        \includegraphics[width=\linewidth]{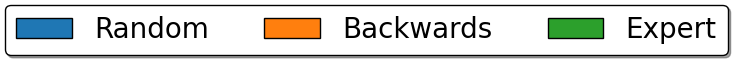}
    \end{subfigure}
    \\
    \begin{subfigure}[t]{0.48\linewidth}
        \includegraphics[width=\linewidth]{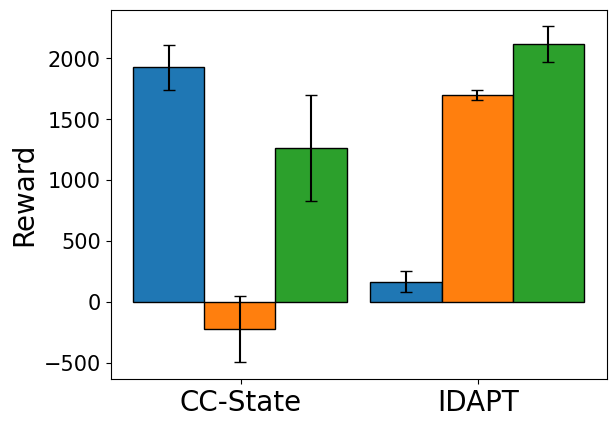}
        \caption{HalfCheetah}
    \end{subfigure}
    \begin{subfigure}[t]{0.48\linewidth}
        \includegraphics[width=\linewidth]{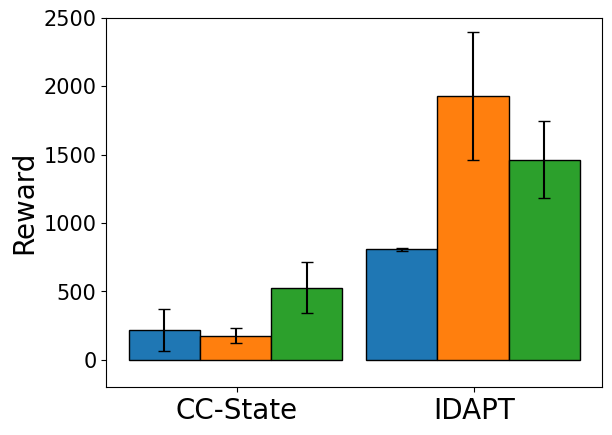}
        \caption{Walker2d}
    \end{subfigure}
    \\
    \vspace{1em}
    \begin{subfigure}[t]{\linewidth}        
        \includegraphics[width=\linewidth]{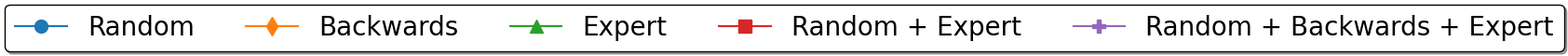}
    \end{subfigure}
    \\
    \begin{subfigure}[t]{0.48\linewidth}
        \includegraphics[width=\linewidth]{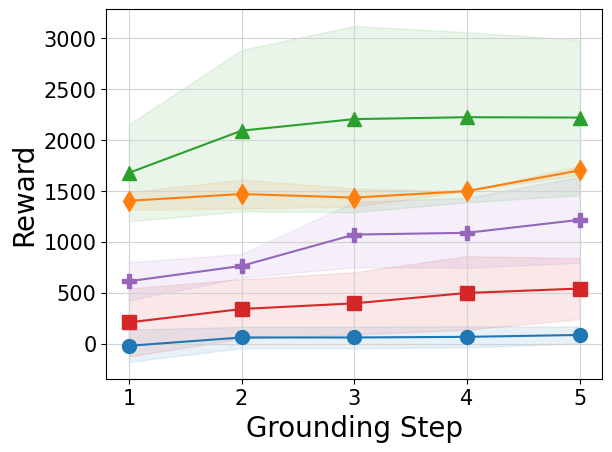}
        \caption{HalfCheetah}
        \label{fig:dataset_exps:halfcheetah}
    \end{subfigure}
        \begin{subfigure}[t]{0.48\linewidth}
        \includegraphics[width=\linewidth]{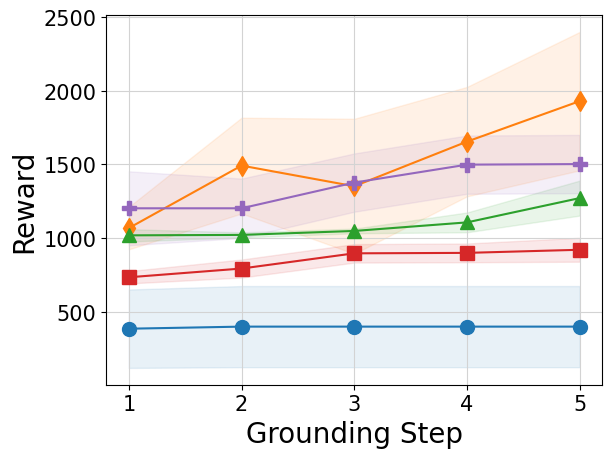}
        \caption{Walker2d}
        \label{fig:dataset_exps:walker2d}
    \end{subfigure}
    \caption{
        Performance of policies transferred to the target domain using initial datasets of different qualities.  ``Random'' is collected by a random agent, ``Backwards" is collected by an agent walking backwards, and ``Expert" is collected by an expert agent on the current task. We compare performances of our method and CC-State (top) and performances of our method over grounding steps (bottom) on different datasets.
    }
    \label{fig:dataset_exps}
\end{figure}

\subsection{Effect of Initial Dataset Quality}
\label{sec:dataset_analysis}

One critical factor for successful domain adaption is the quality of the data used to train transformations or correspondences, specifically its relevance to the current task.  If the dataset does not align with the state distribution or behaviors of the current task, the learned correspondences may not generalize, and therefore fail to transfer a policy for the task.  We analyze the effect of dataset quality on the HalfCheetah and Walker2d tasks for IDAPT and unmodified CC-State (without iterative training).  We use three different datasets, collected by a random policy (Random), a policy trained to move backwards (Backwards), and a policy trained on the current task (Expert).  Here, the Random and Backwards datasets do not align well with the task.  In all experiments, we use datasets of 20k images in both domains and 5 grounding steps and CC-State uses 50k images in both domains following its original experimental setup~\citep{zhang2021learning}.  We report the final transfer performance in terms of reward in \myfigref{fig:dataset_exps}. 

With the exception of HalfCheetah Random data, IDAPT consistently achieves better transfer than CC-State.  With our iterative training procedure we can obtain better task-aligned data through training, which mitigates the effects of a poor initial dataset.  In Walker2d, IDAPT actually performs better with the Backwards dataset rather than the perfectly aligned Expert data.  We hypothesize that this is due to the expert policy in this task having little variability in its behavior, resulting in an image translation that does not generalize well to an imperfect agent, making policy training harder.  Meanwhile, the backwards policy generated a wide range of poses resulting in a more robust image translation for training that can be later improved. While alignment of both datasets with the current task and the breadth of its distribution are key factors that affect the quality of a learned transformation, through iterative grounding, IDAPT is able to partially overcome this issue.

We further examine the use of mixed datasets, ``Random+Expert" and ``Random+Backwards+Expert", consisting of trajectories gathered from multiple different policies.  In \myfigref{fig:dataset_exps:halfcheetah} and \myfigref{fig:dataset_exps:walker2d}, it is clear that the choice of dataset impacts transfer performance. Notably, the Random dataset performs worst and is unable to improve over multiple grounding steps.  Meanwhile, the ``Random+Backwards+Expert'' dataset generally performs well, suggesting that in practice, a mixture dataset of many different behaviors will likely perform well even if some of those behaviors on their own would not result in good transformations.

\subsection{Ablation on Action Transformation}

To verify the importance of closing dynamics domain gaps, we compare the performance of IDAPT with and without the action transformation.  In the ablated model, we only train the visual transformation during the grounding step and apply actions directly in each environment.  \myfigref{fig:result_step} shows a large drop in performance compared to our full model, which quantifies the performance gains by correcting the dynamics gap. Furthermore, with the dynamics domain gap, the distribution of the online target environment data can be far from the source environment data.  When this happens, the quality of the translation will not improve and can lead to worse transfer performance when dynamics differences are unaccounted for.  Thus, it is vital that IDAPT addresses both visual and dynamics domain gaps for the full benefits of iterative grounding.

\subsection{State Reconstruction Regularization for Visual Transformation Training}

We ablate the state reconstruction regularization in our visual transformation model.  We train IDAPT with and without the state reconstruction regularization and compare performance on the target environment with only the visual domain gap during the initial policy training phase. The results demonstrate that in general, our model with state reconstruction regularization achieves better transfer performance (see appendix, \myfigref{fig:norecon_ablation}).

We further perform the same ablation on target-hard environments over multiple grounding steps. The results in appendix, \myfigref{fig:norecon_grounding_ablation} show that the state reconstruction regularization improves performance significantly over all environments.  Qualitatively, we observe that the regularization helps minimize artifacts and stabilize CycleGAN training, resulting in better visual transformations.

\section{Conclusion}

We propose IDAPT, a novel policy transfer approach that addresses visual and dynamics domain gaps with minimal assumptions in the target environment by grounding the source environment in the target environment with a visual transformation and action transformation between domains.  IDAPT iteratively updates the transformations and optimizes a policy in the grounded environment to progressively align the domains and train a policy.  We demonstrate that IDAPT outperforms domain randomization methods which struggle to learn in high randomization regimes.
IDAPT is designed for target environments that have difficulties in collecting task supervision and interactions, which is not limited to sim-to-sim transfer. Therefore, applying IDAPT to sim-to-real and real-to-real policy transfer is a promising future direction.

\section*{Acknowledgments}
This research is supported by the Annenberg Fellowship from USC. We would like to thank the USC CLVR lab members for constructive feedback.

\bibliographystyle{plainnat}
\bibliography{main}

\clearpage
\appendix
\section{Appendix}

\subsection{Results on Normalized Reward}

\begin{figure}[ht]
    \centering
    \begin{subfigure}[t]{\linewidth}
        \centering
        \includegraphics[width=\linewidth]{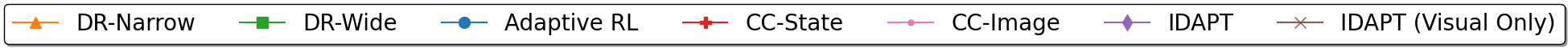}
    \end{subfigure}
    \\
    \begin{subfigure}[t]{0.45\linewidth}
        \centering
        \includegraphics[width=\linewidth]{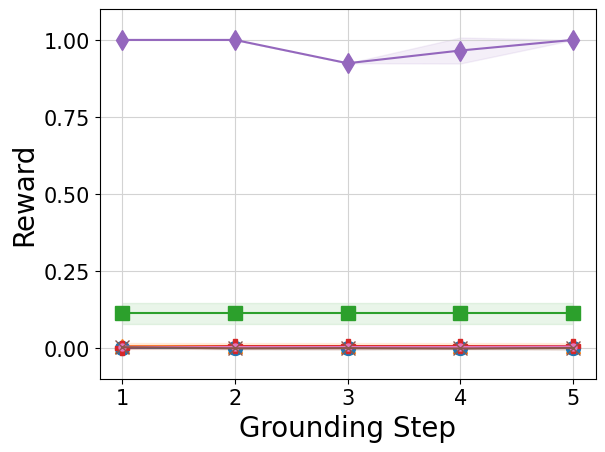}
        \caption{InvertedPendulum}
        \label{fig:result_step_norm:invertedpendulum}
    \end{subfigure}
    \begin{subfigure}[t]{0.45\linewidth}
        \centering
        \includegraphics[width=\linewidth]{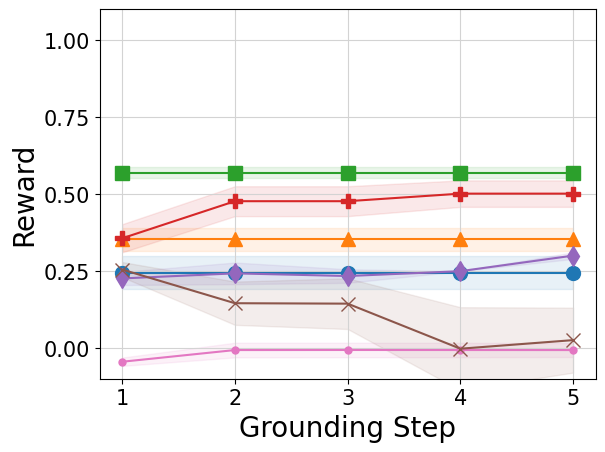}
        \caption{HalfCheetah}
        \label{fig:result_step_norm:halfcheetah}
    \end{subfigure}
    \\
    \begin{subfigure}[t]{0.45\linewidth}
        \centering
        \includegraphics[width=\linewidth]{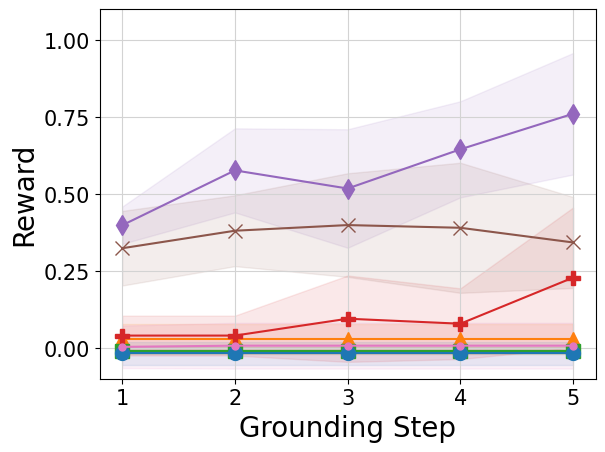}
        \caption{Walker2d}
        \label{fig:result_step_norm:walker2d}
    \end{subfigure}
    \begin{subfigure}[t]{0.45\linewidth}
        \centering
        \includegraphics[width=\linewidth]{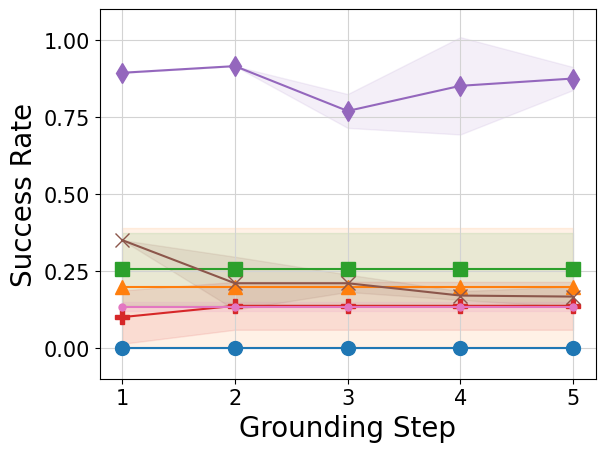}
        \caption{Fetch-Reach}
        \label{fig:result_step_norm:fetchreach}
    \end{subfigure}
    \\
    \begin{subfigure}[t]{0.45\linewidth}
        \centering
        \includegraphics[width=\linewidth]{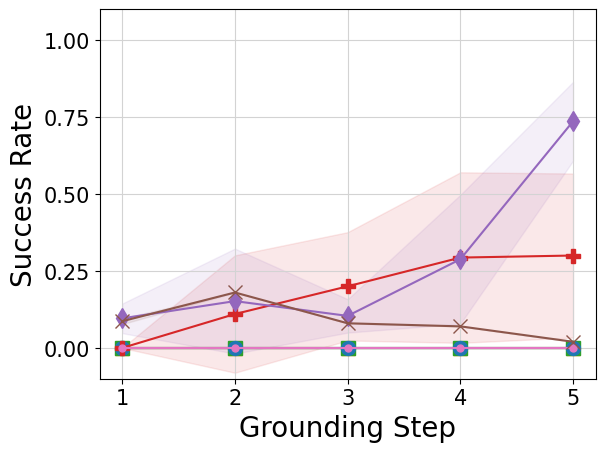}
        \caption{Sawyer-Push}
        \label{fig:result_step_norm:sawyer_push}
    \end{subfigure}
    \caption{
        Comparisons of all methods on the target-hard task with normalized rewards.  The lower bound is the performance of a policy trained in the source environment, then evaluated in the target.  The upper bound is a policy training directly in the target environment.
    }
\label{fig:result_step_norm}
\end{figure}

In addition to \myfigref{fig:result_step}, we report the normalized results for the experiments in \myfigref{fig:result_step_norm}.  For each task, we normalize the reward or success rate between the target environment performance of a policy trained in the source environment (lower bound) and target environment (upper bound).  In every task except HalfCheetah, IDAPT results in more than 50\% of the optimal performance.

\subsection{Additional Analysis}

\subsubsection{\textbf{Data Accumulation}}

\begin{figure}[ht]
    \centering
    \begin{subfigure}[t]{0.7\linewidth}        
        \includegraphics[width=\linewidth]{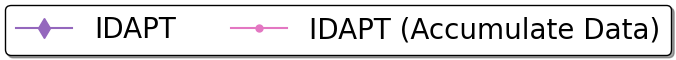}
    \end{subfigure}
    \\
    \begin{subfigure}[t]{0.48\linewidth}
        \includegraphics[width=\linewidth]{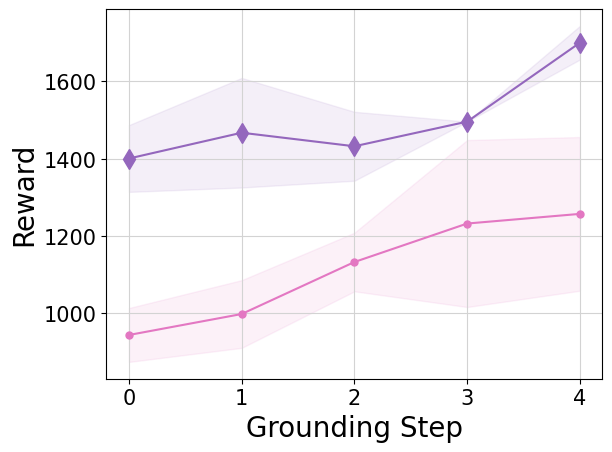}
        \caption{HalfCheetah}
    \end{subfigure}
    \begin{subfigure}[t]{0.48\linewidth}
        \includegraphics[width=\linewidth]{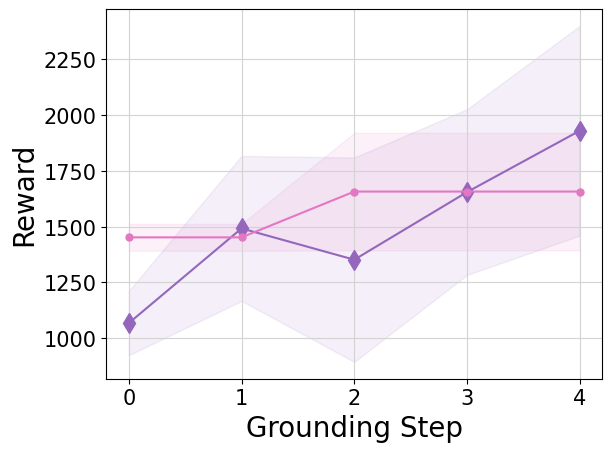}
        \caption{Walker2d}
    \end{subfigure}
    \caption{
        Performance of policies transferred to the target domain using different data accumulation options.  For \method~(Accumulate Data), we used double the number of finetuning epochs.
    }
\label{fig:accumulate_data}
\end{figure}

For each grounding step, we chose to use the online dataset of 1k interactions to finetune the visual transformation model.  An alternative would be to use the accumulated data gathered at each grounding step along with the large task agnostic dataset to make use of all available data.  We compare the two in \myfigref{fig:accumulate_data} and find that there is no performance improvement with data accumulation.  In fact, we may have to train for significantly more epochs or use a data balancing scheme to achieve similar results.  We hypothesize that the use of a small number of finetuning epochs and the fact that the most recent online dataset is better aligned with the task mitigates any issues with forgetting or overfitting and is more computationally efficient.

\begin{figure}[ht]
    \centering
    \begin{subfigure}[t]{0.5\linewidth}
        \centering
        \includegraphics[width=\linewidth]{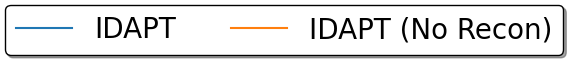}
    \end{subfigure}
    \\
    \begin{subfigure}[t]{0.45\linewidth}
        \centering
        \includegraphics[width=\linewidth]{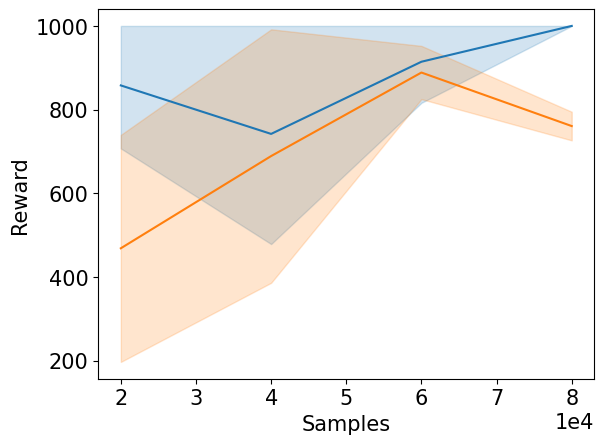}
        \caption{InvertedPendulum}
    \end{subfigure}
    \begin{subfigure}[t]{0.45\linewidth}
        \centering
        \includegraphics[width=\linewidth]{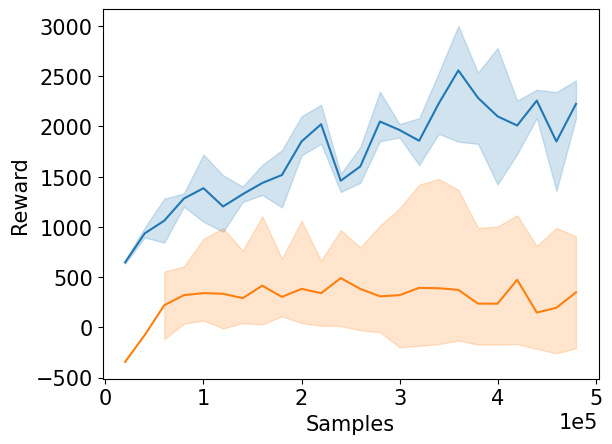}
        \caption{HalfCheetah}
\    \end{subfigure}
    \\
    \begin{subfigure}[t]{0.45\linewidth}
        \centering
        \includegraphics[width=\linewidth]{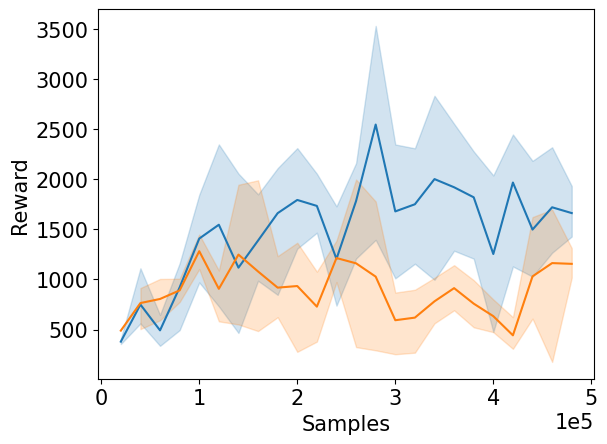}
        \caption{Walker2d}
    \end{subfigure}
    \begin{subfigure}[t]{0.45\linewidth}
        \centering
        \includegraphics[width=\linewidth]{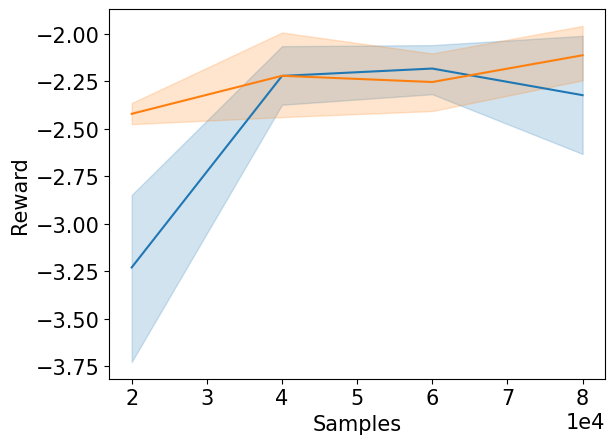}
        \caption{Fetch-Reach}
    \end{subfigure}
    \\
    \begin{subfigure}[t]{0.45\linewidth}
        \centering
        \includegraphics[width=\linewidth]{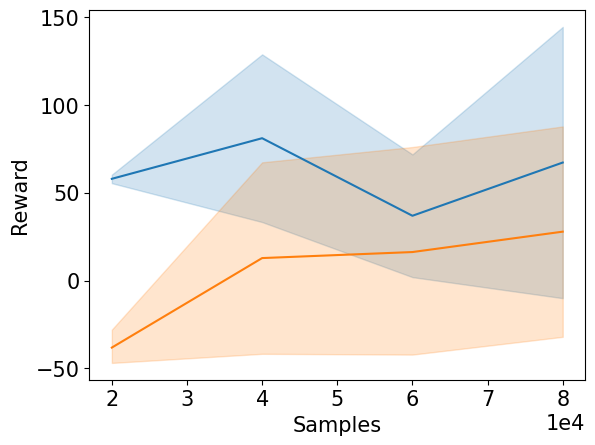}
        \caption{Sawyer-Push}
    \end{subfigure}
    \caption{
        Learning curve of policy in the target visual environment, without dynamics differences, throughout policy training in source environment to convergence.  
    }
\label{fig:norecon_ablation}
\end{figure}

\begin{figure}[ht]
    \centering
    \begin{subfigure}[t]{0.5\linewidth}
        \centering
        \includegraphics[width=\linewidth]{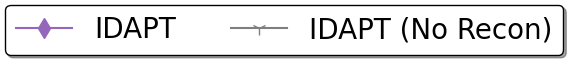}
    \end{subfigure}
    \\
    \begin{subfigure}[t]{0.45\linewidth}
        \centering
        \includegraphics[width=\linewidth]{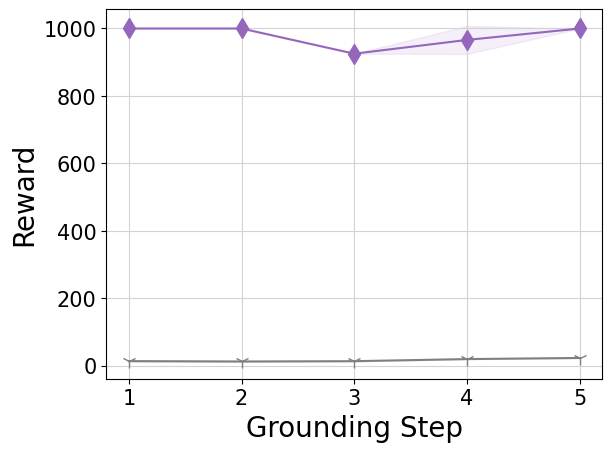}
        \caption{InvertedPendulum}
    \end{subfigure}
    \begin{subfigure}[t]{0.45\linewidth}
        \centering
        \includegraphics[width=\linewidth]{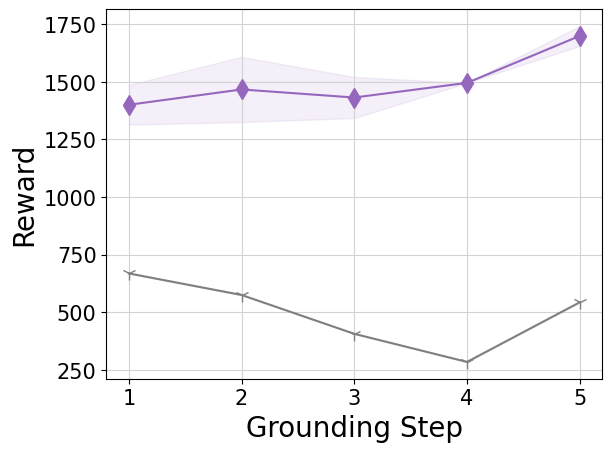}
        \caption{HalfCheetah}
\    \end{subfigure}
    \\
    \begin{subfigure}[t]{0.45\linewidth}
        \centering
        \includegraphics[width=\linewidth]{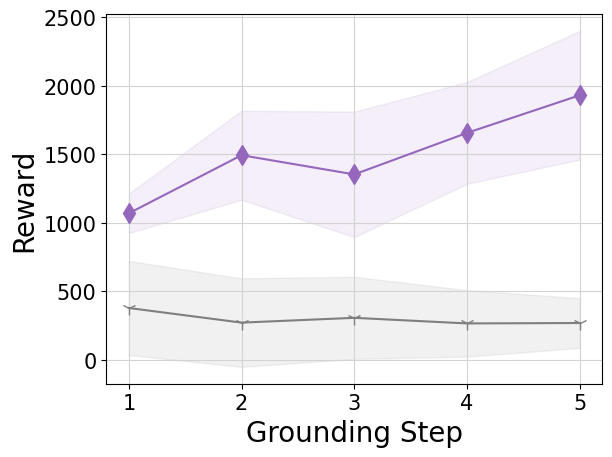}
        \caption{Walker2d}
    \end{subfigure}
    \begin{subfigure}[t]{0.45\linewidth}
        \centering
        \includegraphics[width=\linewidth]{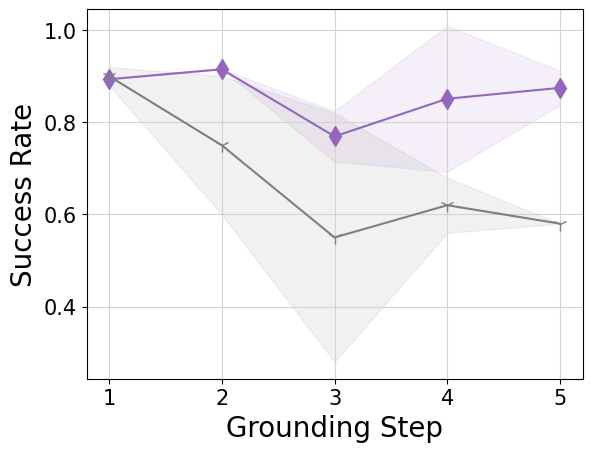}
        \caption{Fetch-Reach}
    \end{subfigure}
    \\
    \begin{subfigure}[t]{0.45\linewidth}
        \centering
        \includegraphics[width=\linewidth]{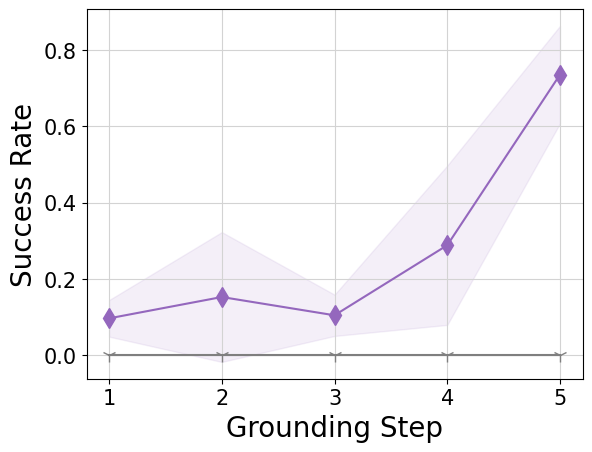}
        \caption{Sawyer-Push}
    \end{subfigure}    
    \caption{
    Policy transfer performance in target-hard environments with and without state reconstruction regularization. The results show that the state reconstruction regularization is essential for our method to learn the visual transformation.
    }
\label{fig:norecon_grounding_ablation}
\end{figure}

\subsubsection{\textbf{Additional Datasets Analysis}}

\begin{figure}[ht]
    \centering
    \begin{subfigure}[t]{\linewidth}        
        \includegraphics[width=\linewidth]{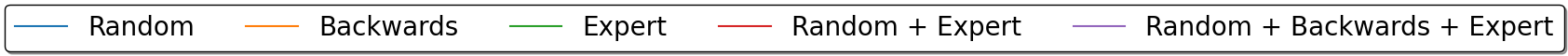}
    \end{subfigure}
    \\
    \begin{subfigure}[t]{0.48\linewidth}
        \includegraphics[width=\linewidth]{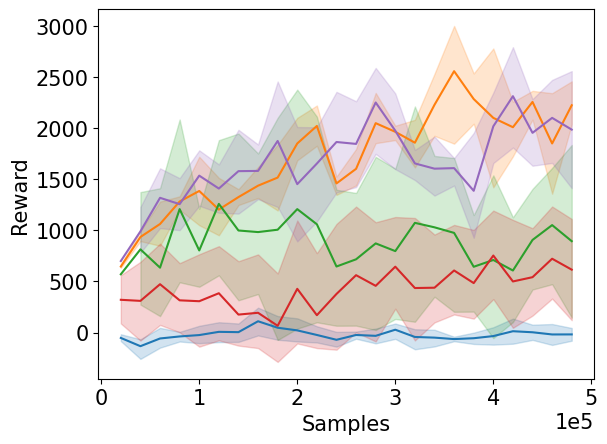}
        \caption{HalfCheetah}
    \end{subfigure}
    \begin{subfigure}[t]{0.48\linewidth}
        \includegraphics[width=\linewidth]{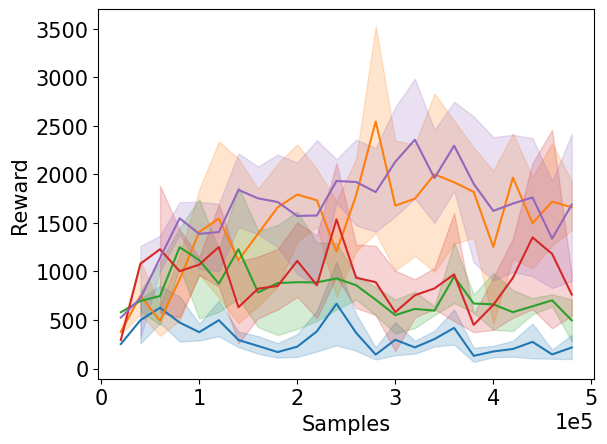}
        \caption{Walker2d}
    \end{subfigure}
    \caption{
        Learning curve of policy in the target visual environment, without dynamics differences, throughout policy training in source environment to convergence.  We use different initial datasets composed of trajectories collected by a random policy, a policy trained to walk backwards, and the expert policy.  
    }
\label{fig:datasets_ablations}
\end{figure}

The quality of the task-agnostic dataset can affect the initial grounding of the training environment.  We investigate the effect of using datasets gathered by different policies (Random, Backwards, Expert) in HalfCheetah and Walker2d tasks.  In addition, we also use mixed datasets, ``Random+Expert'' and ``Random+Backwards+Expert'', consisting of trajectories gathered from multiple different policies.  We look at both the first policy training step in a target environment with visual domain gap only (\myfigref{fig:datasets_ablations}) is isolate the visual domain difference and performance over multiple grounding steps in the target-hard environment (\myfigref{fig:dataset_exps:halfcheetah} and \myfigref{fig:dataset_exps:walker2d}).  In both cases, it is clear that the choice of dataset impacts transfer performance.  Notably, the Random dataset performs worst and is unable to improve over multiple grounding steps.  Meanwhile, the ``Random+Backwards+Expert'' dataset performs well, suggesting that in practice, a mixture dataset of many different behaviors will likely perform well even if some of those behaviors on their own would not result in good transformations.  Furthermore, we see that for most datasets, the performance does improve over multiple grounding steps, which allows IDAPT to partially overcome poor initial transformations.

\subsection{Quality of Visual Transformations}

\begin{figure}[ht]
    \centering
    \begin{subfigure}[t]{\linewidth}
        \centering
        \makebox[0.2\linewidth]{Source}
        \makebox[0.2\linewidth]{Translated}
        \makebox[0.2\linewidth]{Target}
        \makebox[0.2\linewidth]{Diff}
    \end{subfigure}
    \\
    \begin{subfigure}{\linewidth}
        \centering
        \includegraphics[width=0.8\linewidth]{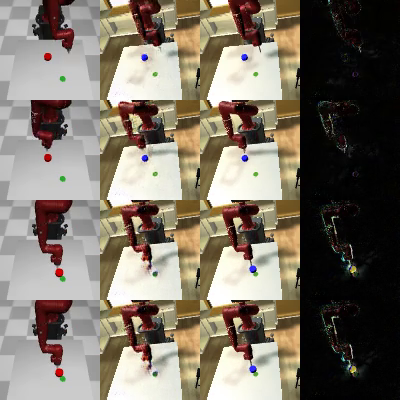}
        \caption{Before grounding step}
    \end{subfigure} 
    \\
    \begin{subfigure}{\linewidth}
        \centering
        \includegraphics[width=0.8\linewidth]{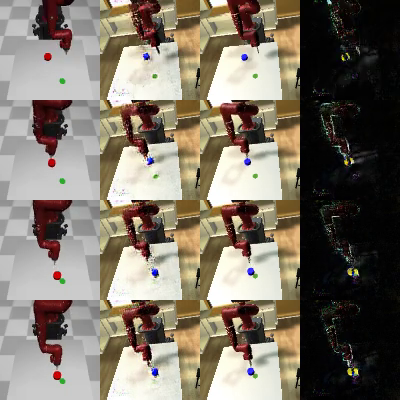}
        \caption{After grounding step}
    \end{subfigure}
    \caption{
        Translated images across visual domains for the Sawyer-Push task using the visual transformation after pretraining (a) and after finetuning with one grounding step (b).  
        Top row of each section is a series of source environment images, bottom row is the corresponding translated target environment image.  
    }
    \label{fig:gen_imgs}
\end{figure}

To evaluate the visual transformation quality, we show pairs of images from target and source domains that share the \textit{same underlying state}, and compare the difference between images generated by the visual transformation with source domain images and true target domain images, in \myfigref{fig:gen_imgs}. Before grounding, the puck disappears in the later frames because the Sawyer rarely moves the puck in the pretraining dataset.  After grounding, our visual transformation learns to translate the puck position correctly. 

\begin{figure}[ht]
    \centering
    \begin{subfigure}[t]{\linewidth}
        \centering
        \makebox[0.2\linewidth]{Source}
        \makebox[0.2\linewidth]{Translated}
        \makebox[0.2\linewidth]{Target}
        \makebox[0.2\linewidth]{Diff}
    \end{subfigure}
    \\
    \begin{subfigure}{\linewidth}
        \centering
        \includegraphics[width=0.8\linewidth]{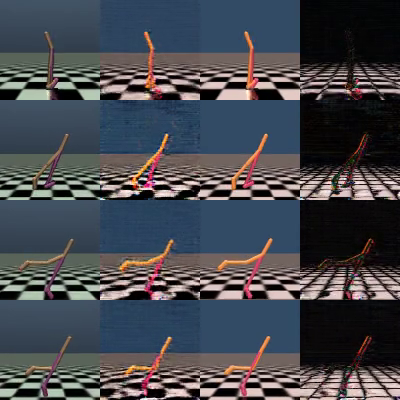}
    \end{subfigure}
    \caption{
        Translated images across visual domains for the Walker2D target-easy task using the visual transformation.  
    }
    \label{fig:walker_flipped_floor}
\end{figure}

In HalfCheetah and Walker2D, we noticed that occasionally the colors of the checkered floor flipped in the translated image, resulting in large pixel-wise errors (\myfigref{fig:walker_flipped_floor}).  The repeated floor pattern and the gait of the backwards agent used to gather the initial data result in a uniform floor distribution that makes it difficult for the CycleGAN to learn the correct alignment.  However, this did not affect the transfer performance of the policy, demonstrating that the learned policy can be robust to errors in translation that are task irrelevant.  This is possibly due to an invariance introduced by the random cropping image augmentation during training.  

\subsection{Environment Details}
\label{sec:environment_details}

\subsubsection{\textbf{Locomotion Environments}}
The locomotion environments are modified from OpenAI Gym~\citep{brockman2016openai} and use the same default action space and visual domain. We created the target-easy visual domains by varying lighting and background color.  For the target-hard visual domain, we additionally varied background texture, character color, and viewpoint.

\subsubsection{\textbf{Manipulation Environments}}
Our manipulation environments included Fetch-Reach which is modified form OpenAI Gym, and Sawyer-Push which uses a simulation of he 7-DoF Rethink Sawyer. To create the dynamics of Fetch-Reach target environment we rotate the action vector around z-axis and add a bias to the third coordinate of the action vector. For Sawyer, we vary the friction and mass to create target environment dynamics. For visual of the target domain, we change colors,  lighting, and viewpoint in easy target environment, and we use Unity3D rendering with realistic lighting and background in the hard target environment.

\subsection{Baseline Implementations}
\label{sec:baseline_implementation}


\subsubsection{\textbf{Domain Randomization Implementation}}
\label{sec:dr_implementation}
\begin{figure}[ht]
    \centering
    \begin{subfigure}[t]{0.7\linewidth}
        \centering
        \includegraphics[width=\linewidth]{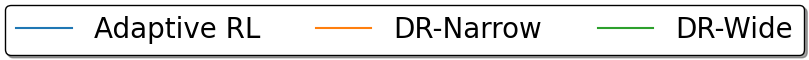}
    \end{subfigure}
    \\
    \begin{subfigure}[t]{0.45\linewidth}
        \centering
        \includegraphics[width=\linewidth]{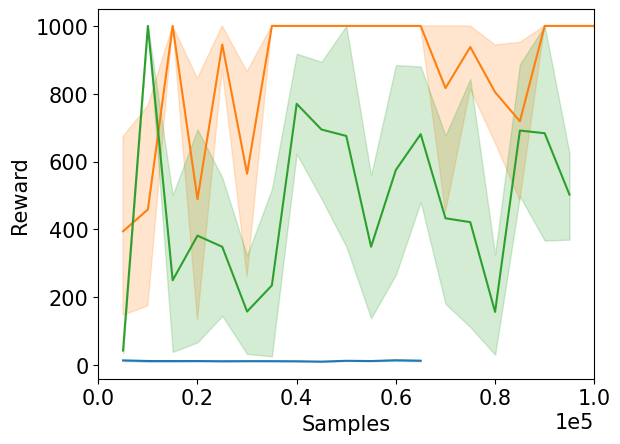}
        \caption{InvertedPendulum}
        \label{fig:baseline_learning_curves:invertedpendulum}
    \end{subfigure}
    \begin{subfigure}[t]{0.45\linewidth}
        \centering
        \includegraphics[width=\linewidth]{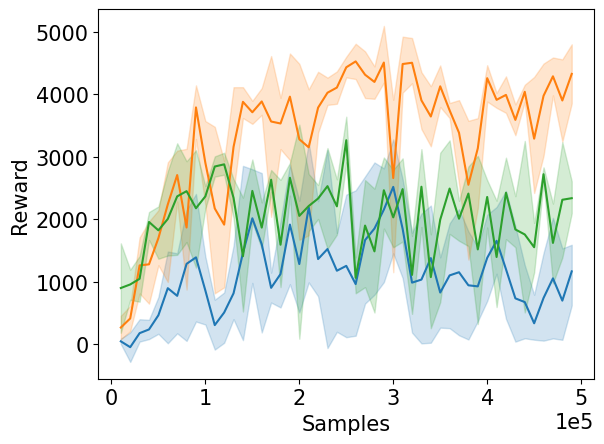}
        \caption{HalfCheetah}
        \label{fig:baseline_learning_curves:halfcheetah}
    \end{subfigure}
    \\
    \begin{subfigure}[t]{0.45\linewidth}
        \centering
        \includegraphics[width=\linewidth]{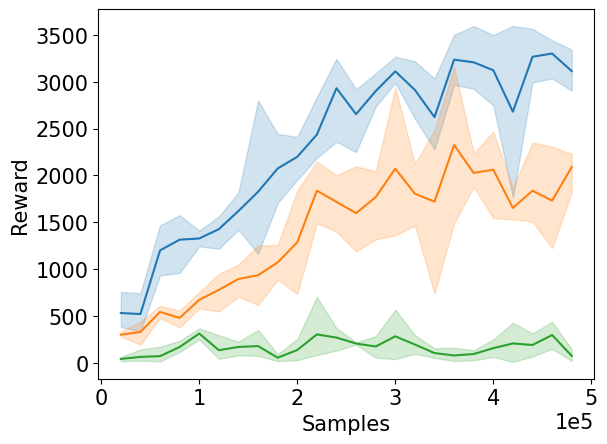}
        \caption{Walker2d}
        \label{fig:baseline_learning_curves:walker2d}
    \end{subfigure}
    \begin{subfigure}[t]{0.45\linewidth}
        \centering
        \includegraphics[width=\linewidth]{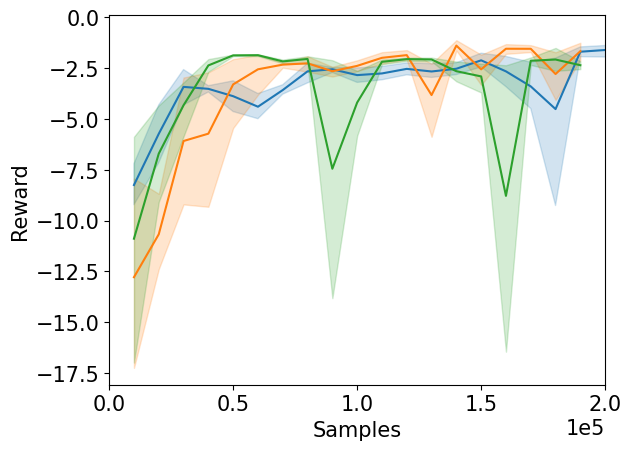}
        \caption{Fetch-Reach}
        \label{fig:baseline_learning_curves:fetchreach}
    \end{subfigure}
    \\
    \begin{subfigure}[t]{0.45\linewidth}
        \centering
        \includegraphics[width=\linewidth]{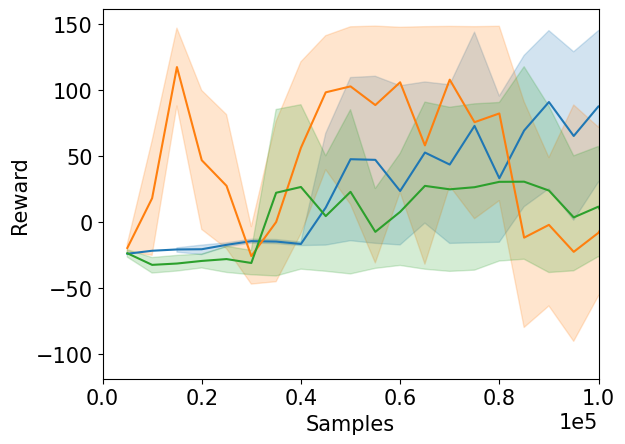}
        \caption{Sawyer-Push}
        \label{fig:baseline_learning_curves:sawyer_push}
    \end{subfigure}
    \caption{
        Learning curves of baseline methods on training environments.
    }
\label{fig:baseline_learning_curves}
\end{figure}

\begin{figure}[t]
    \centering
    \begin{subfigure}[t]{\linewidth}
        \centering
        \includegraphics[width=0.3\linewidth]{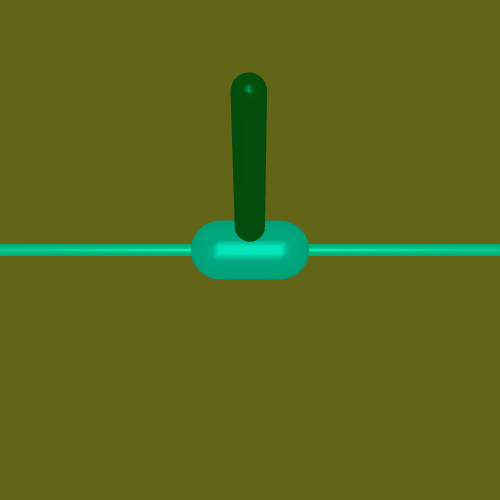}
        \includegraphics[width=0.3\linewidth]{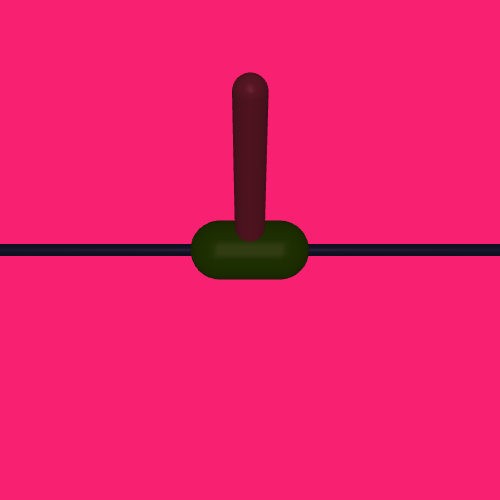}
        \includegraphics[width=0.3\linewidth]{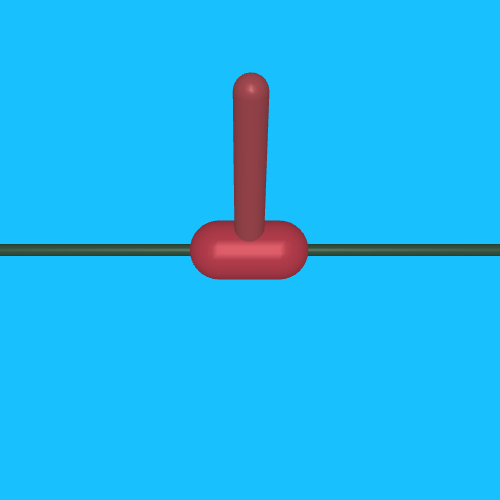}
        \caption{InvertedPendulum}
    \end{subfigure}
    \\
    \begin{subfigure}[t]{\linewidth}
        \centering
        \includegraphics[width=0.3\linewidth]{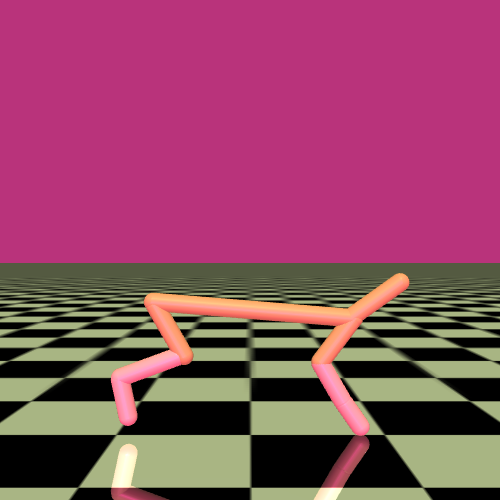}
        \includegraphics[width=0.3\linewidth]{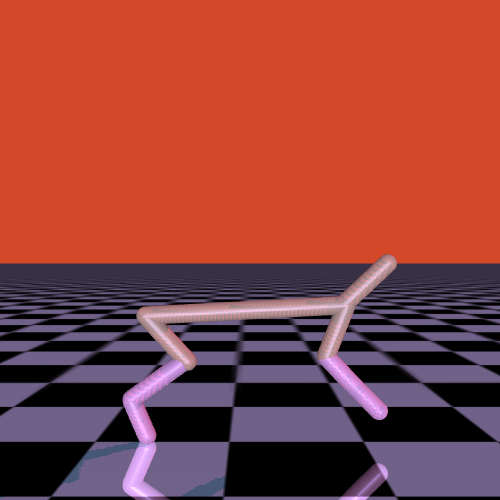}
        \includegraphics[width=0.3\linewidth]{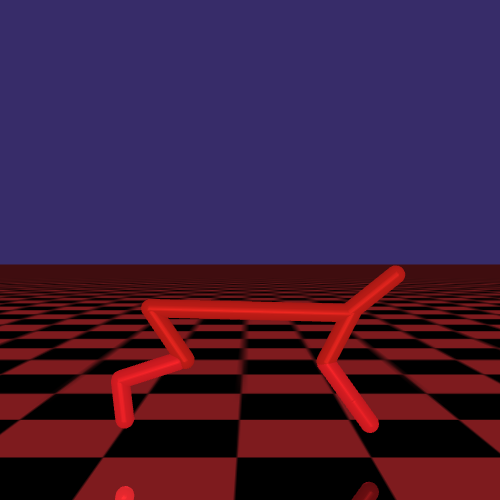}
        \caption{HalfCheetah}
    \end{subfigure}
    \\
    \begin{subfigure}[t]{\linewidth}
        \centering
        \includegraphics[width=0.3\linewidth]{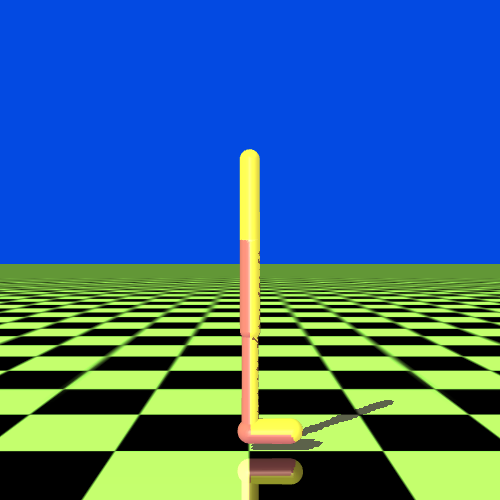}
        \includegraphics[width=0.3\linewidth]{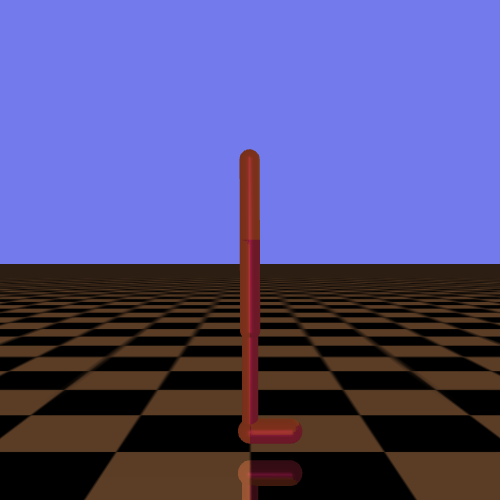}
        \includegraphics[width=0.3\linewidth]{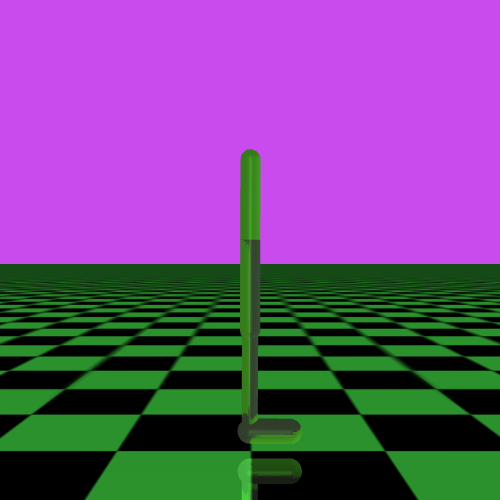}
        \caption{Walker2d}
    \end{subfigure}
    \\
    \begin{subfigure}[t]{\linewidth}
        \centering
        \includegraphics[width=0.3\linewidth]{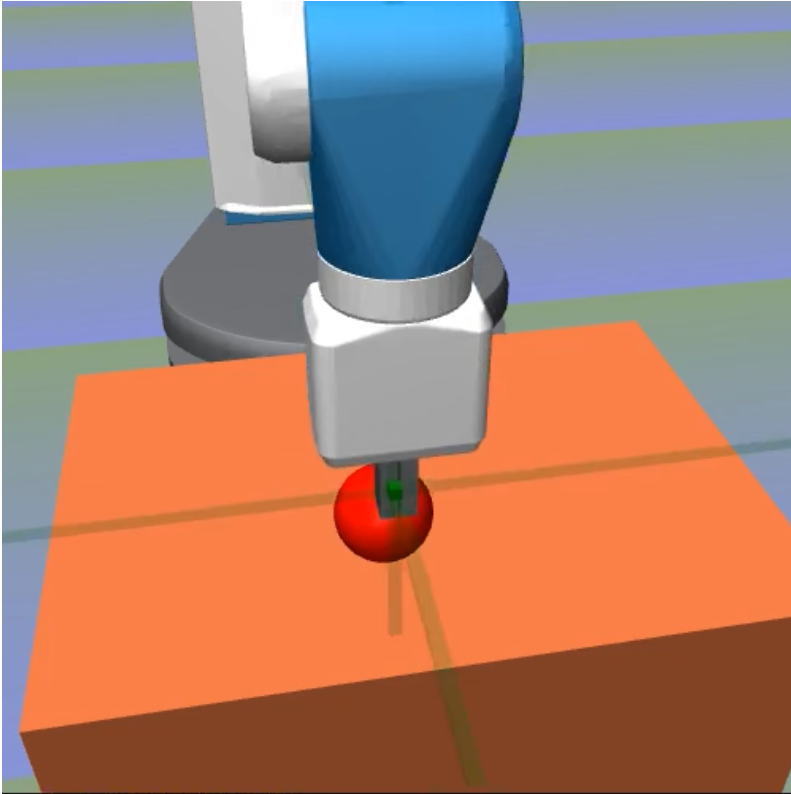}
        \includegraphics[width=0.3\linewidth]{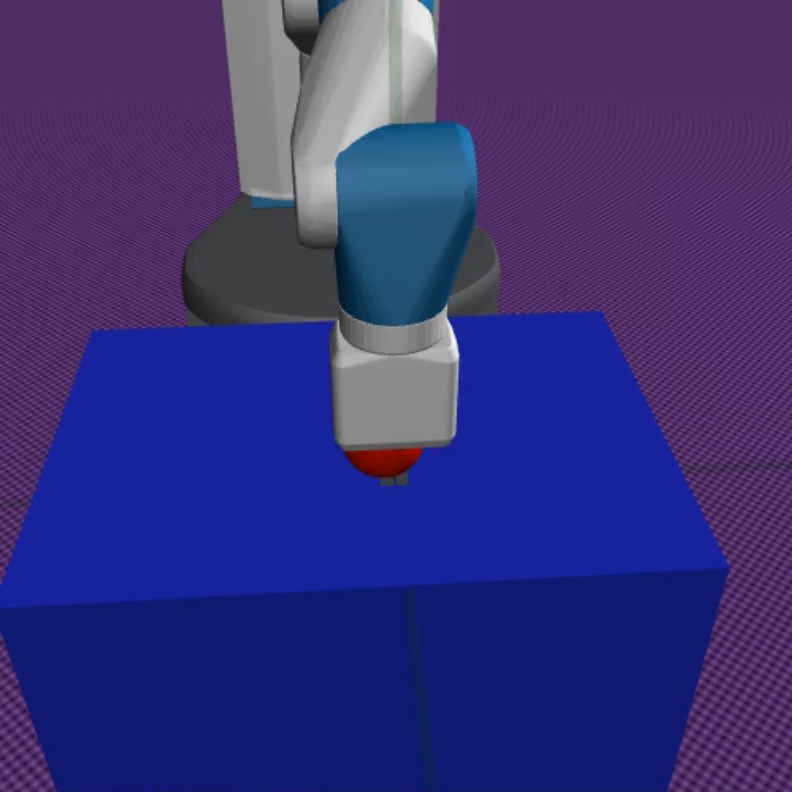}
        \includegraphics[width=0.3\linewidth]{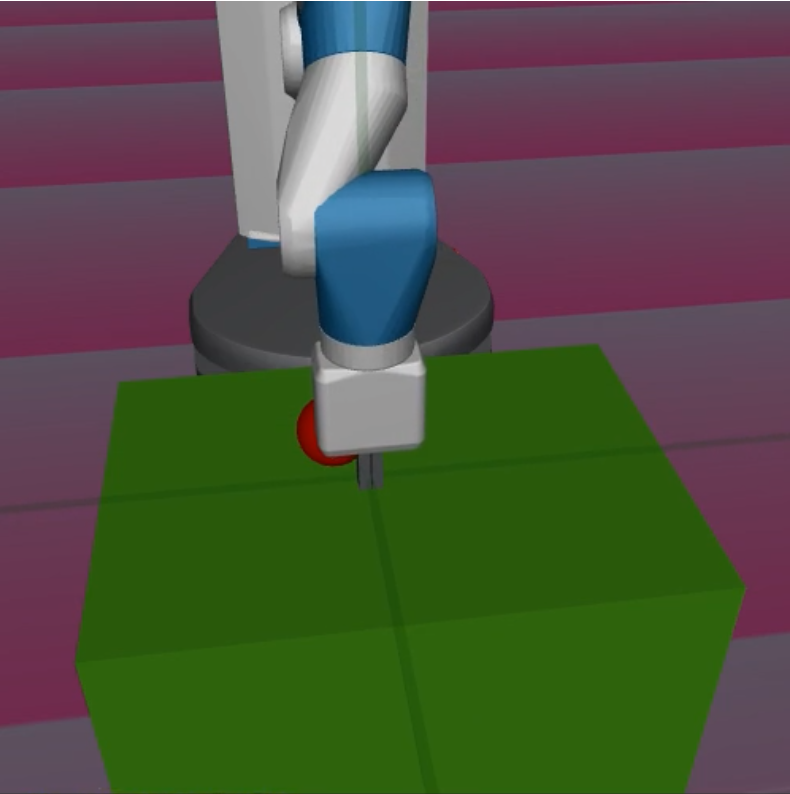}
        \caption{Fetch-Reach}
    \end{subfigure}
    \\
    \begin{subfigure}[t]{\linewidth}
        \centering
        \includegraphics[width=0.3\linewidth]{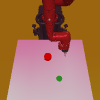}
        \includegraphics[width=0.3\linewidth]{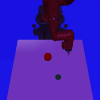}
        \includegraphics[width=0.3\linewidth]{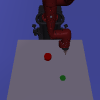}
        \caption{Sawyer-Push}
    \end{subfigure}
    \caption{
        Narrow range visualize domain randomization examples, including color and lighting changes.
    }
    \label{fig:dr_visual_narrow}
\end{figure}

\begin{figure}[t]
    \centering
    \begin{subfigure}[t]{\linewidth}
        \centering
        \includegraphics[width=0.3\linewidth]{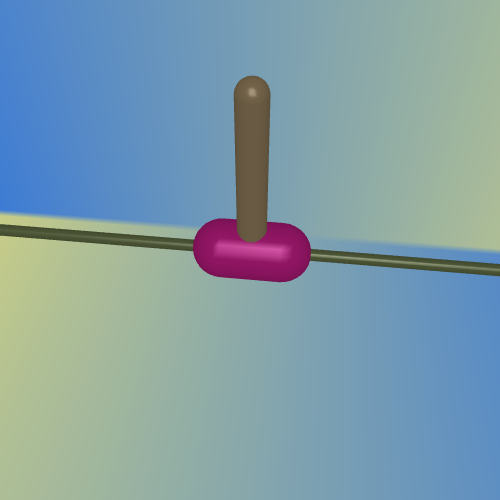}
        \includegraphics[width=0.3\linewidth]{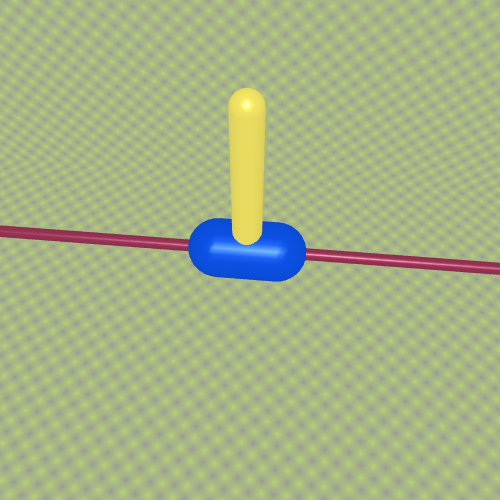}
        \includegraphics[width=0.3\linewidth]{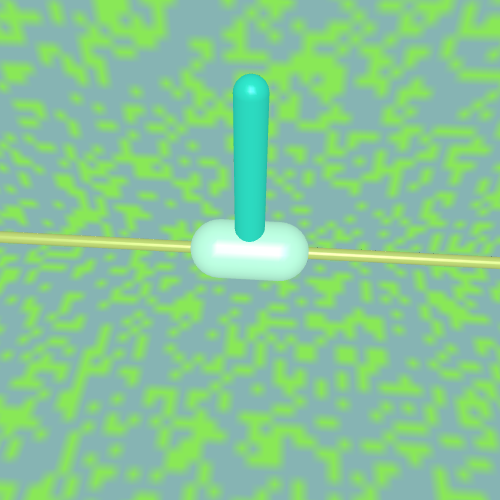}
        \caption{InvertedPendulum}
    \end{subfigure}
    \\
    \begin{subfigure}[t]{\linewidth}
        \centering
        \includegraphics[width=0.3\linewidth]{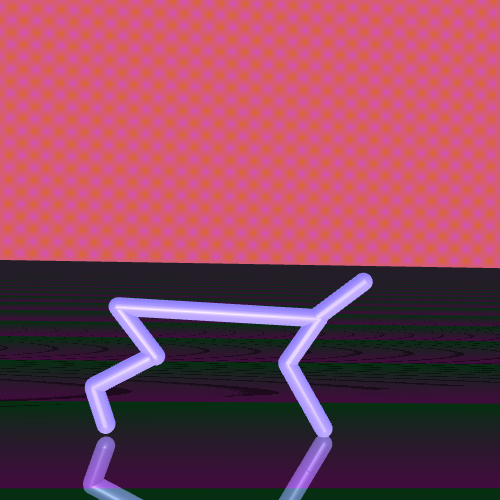}
        \includegraphics[width=0.3\linewidth]{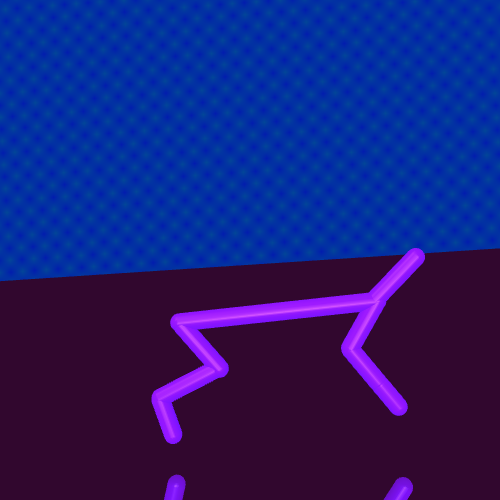}
        \includegraphics[width=0.3\linewidth]{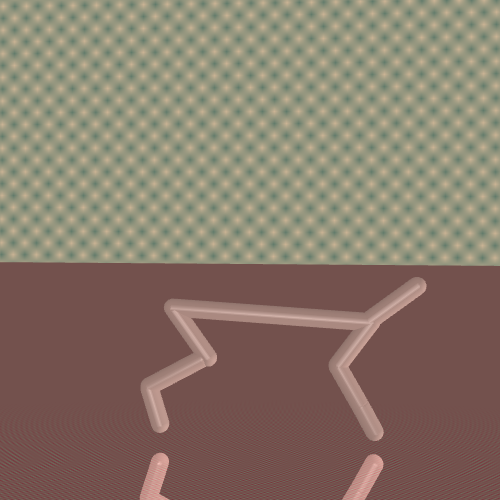}
        \caption{HalfCheetah}
    \end{subfigure}
    \\
    \begin{subfigure}[t]{\linewidth}
        \centering
        \includegraphics[width=0.3\linewidth]{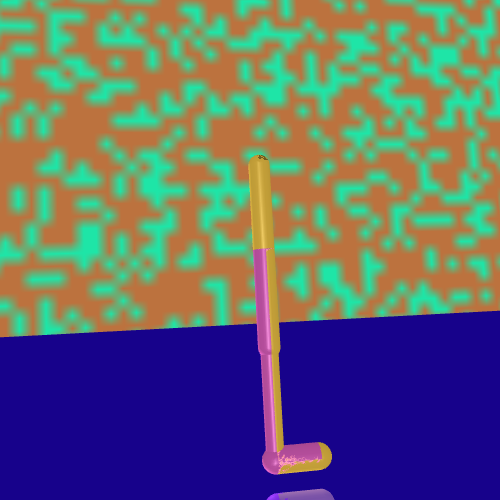}
        \includegraphics[width=0.3\linewidth]{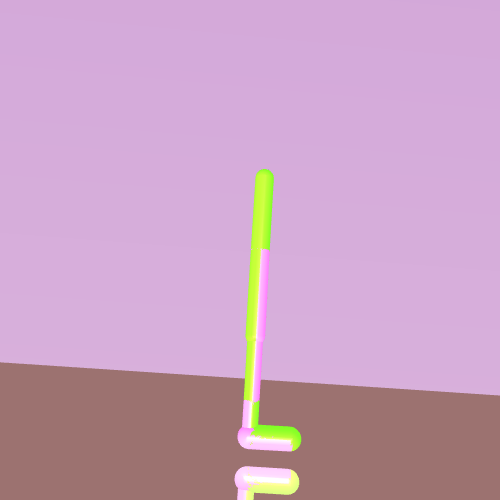}
        \includegraphics[width=0.3\linewidth]{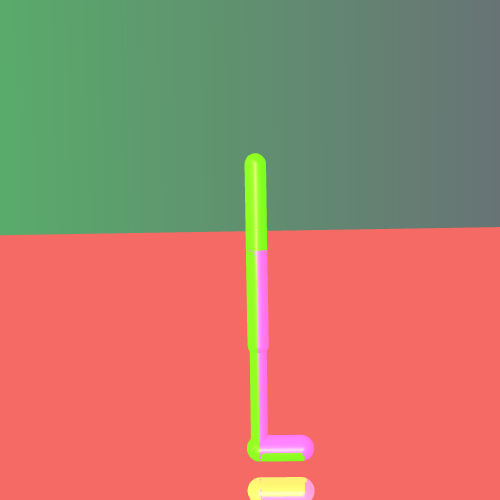}
        \caption{Walker2d}
    \end{subfigure}
    \\
    \begin{subfigure}[t]{\linewidth}
        \centering
        \includegraphics[width=0.3\linewidth]{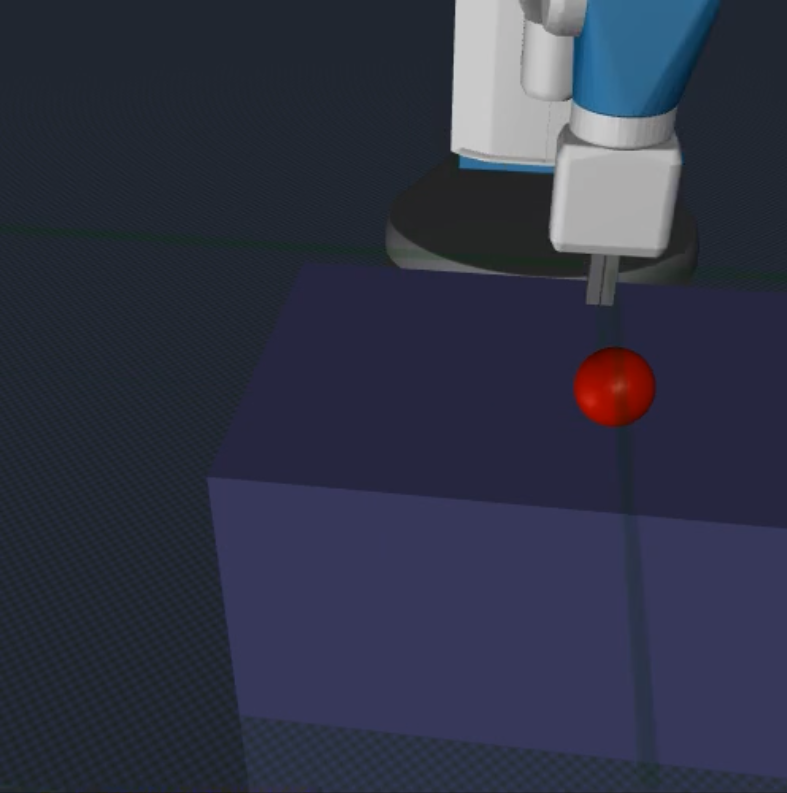}
        \includegraphics[width=0.3\linewidth]{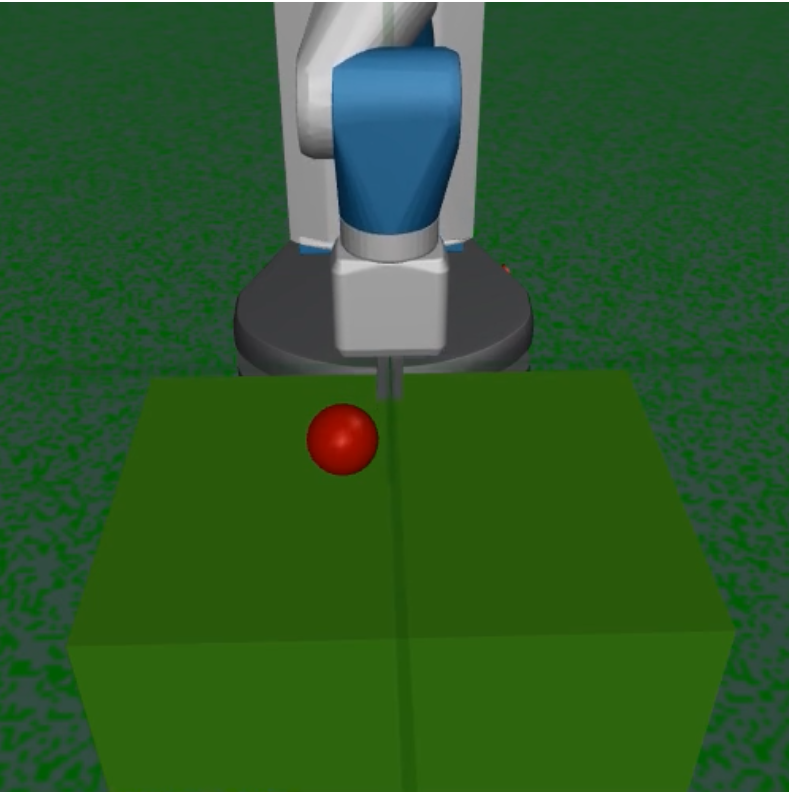}
        \includegraphics[width=0.3\linewidth]{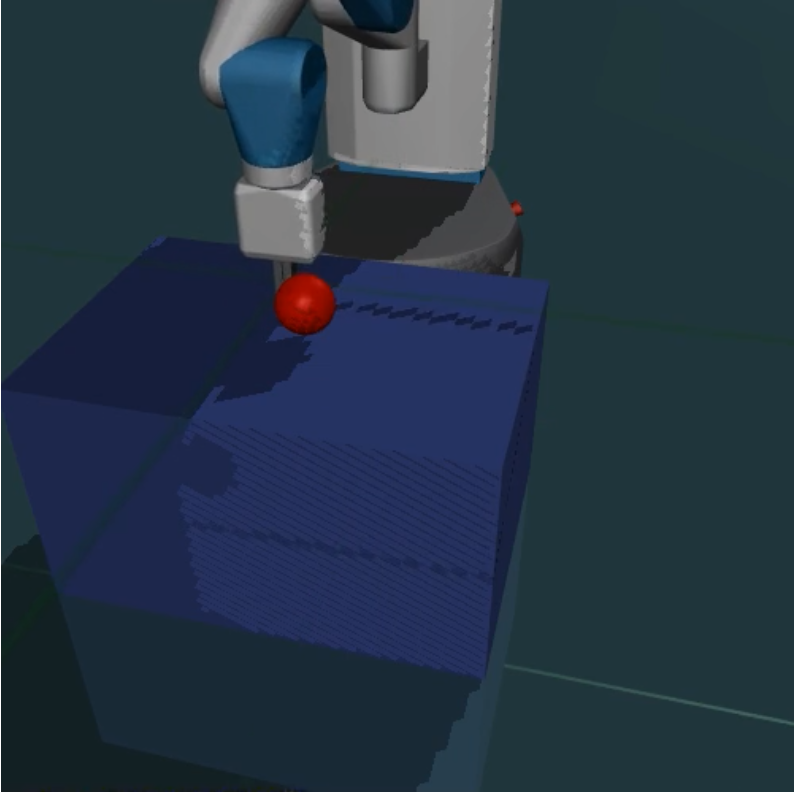}
        \caption{Fetch-Reach}
    \end{subfigure}
    \\
    \begin{subfigure}[t]{\linewidth}
        \centering
        \includegraphics[width=0.3\linewidth]{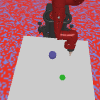}
        \includegraphics[width=0.3\linewidth]{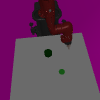}
        \includegraphics[width=0.3\linewidth]{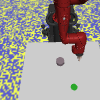}
        \caption{Sawyer-Push}
    \end{subfigure}
    \caption{
        Wide range visualize domain randomization examples, including viewpoint changes and more textures in addition to color and lighting changes.
    }
    \label{fig:dr_visual_wide}
\end{figure}

We implemented domain randomization by modifying simulation parameters every iteration with uniformly sampled random values. The sampling range for dynamics parameters are specified in Table. \ref{tab:dr_dynamics}. Example images of visual randomization we used during training can be found in \myfigref{fig:dr_visual_wide} and \myfigref{fig:dr_visual_narrow}. Additionally we use random cropping from 100x100 pixels images to 92x92 pixels.  We train InvertedPendulum for 5e4 environment steps, HalfCheetah and Walker2D for 5e5 steps, and Sawyer-Push and Fetch-Reach for 1e5 steps. For policy optimization, we use asymmetric SAC~\citep{pinto2017asymmetric} and use the same hyperparameters our method uses~\mytbref{tab:sac_hyperparameter}.  We include learning curves in \myfigref{fig:baseline_learning_curves}.  For some environments, DR-Wide was unable to reach good performance even in the training environment.

\begin{table}[ht]
\centering
\caption{Physics parameters for domain randomization.}
\begin{tabular}{ ccccc } 
 \toprule
 \multirow{2}{7.5em}{\centering Task} & \multirow{2}{6.5em}{\centering Parameter} & \multicolumn{2}{c}{Target} \\ 
 & & Easy & Hard \\
 \midrule
 InvertedPendulum & Pendulum mass & $4 \sim 55$ & $4 \sim 220$ \\ 
 \midrule
 HalfCheetah & Armature & $0.08 \sim 0.25$ & $0.08 \sim 0.44$ \\ 
 \midrule
 Walker2d & Torso mass & $3 \sim 6$ & $3 \sim 11$ \\ 
 \midrule
 \multirow{2}{7.5em}{\centering Fetch-Reach} & Action Rotation & $-30^{\circ}\sim30^{\circ}$ & $-45^{\circ}\sim45^{\circ}$\\
 & Action Bias & $-0.55 \sim 0$ & $-0.55 \sim 0.55$ \\
 \midrule
\multirow{2}{7.5em}{\centering Sawyer-Push} & Puck mass & $0.01 \sim 0.033$ & $0.01 \sim 0.05$ \\ 
& Puck Friction & $ 2 \sim 3.3$ & $2 \sim 4.4$\\
 \bottomrule
\end{tabular}
\label{tab:dr_dynamics}
\end{table}

\subsubsection{\textbf{Adaptive RL Implementation}}
We use an asymmetric SAC agent and use LSTM for both actor and critic network. During training, the agent collects experiences from domain randomized environments using the same domain randomization parameters as DR-Wide. During network updates, we use randomly sampled sequences of 7 time steps to calculate actor and critic loss. To initialize LSTM states during network update, We use stored the internal states of LSTM in both actor and critic collected during episode rollouts. We use the same hyperparameters, training steps, and image cropping as Domain Randomization.  We found that adaptive policies are difficult to train in environments with early termination based on failure, specifically InvertedPendulum and Walker2D.  The flexibility of the adaptive policy is outweighed by the difficulty of training in these tasks and would require further engineering to provide a strong comparison.  We include learning curves in \myfigref{fig:baseline_learning_curves}.

\subsubsection{\textbf{Cross-Domain Correspondence Implementation}}
\label{sec:cc_implementation}
We use the implementation provided by the authors~\citep{zhang2021learning} for cross morphology transfer for \textbf{CC-State}.  This model learns action and state-to-state correspondences to transfer a state-based policy across dynamics differences.  For \textbf{CC-Image}, we modified the existing cross-modality and cross-morphology implementations to replicate the cross-physics-and-modality algorithm to the best of our ability using the same architectures and implementation details.  This model learns action and state-to-image correspondences to transfer a state-based policy across dynamics and modality differences.  To make this algorithm iterative, we gather 1k samples from both environments with the current policy and train the model on the online data for each iteration.

\subsection{IDAPT Implementation Details}
\label{sec:idapt_details}

\subsubsection{\textbf{Policy Training}}
We use Asymmetric SAC~\citep{pinto2017asymmetric} to learn an RL policy.  The input to the actor is a stack of 3 consecutive image frames, originally 100x100 pixels and randomly cropped to 92x92.  The input to the critic is the state.  The actor network consists of a 4-layer CNN encoder with output feature space of dimension 50 and a 2-layer MLP with hidden dimensions of 1024, whose output parameterizes a Gaussian distribution over the action space.  For InvertedPendulum, Sawyer-Push, and Fetch-Reach, we train for 1e4 steps per policy training stage, which takes approximately 20 minutes to train on an  NVIDIA Titan X GPU. For HalfCheetah and Walker2D, we train for 2e5 steps, which takes approximately 3 hours.

\subsubsection{\textbf{Visual Transformation: CycleGAN with Regularization}}
We base our CycleGAN implementation on~\citep{zhu2017unpaired} and use the same hyperparameters and architectures.  Additionally, each state prediction network consists of a 4-layer CNN encoder with output dimension of 50 and a 2-layer MLP with hidden dimensions of size 256.  To train the source domain state prediction network, we use the Adam optimizer~\citep{kingma2014adam} with learning rate 3e-4.  We initialize the target domain state prediction network with the weights of the source domain network and train only the top convolutional layer jointly with the CycleGAN generator networks.  During pretraining, we first train the source domain state prediction network for 40 epochs, then train the CycleGAN + target domain state prediction network for 40 epochs.  During finetuning, we train each network group for 5 epochs. Training time for the initial training is approximately 8 hours and for finetuning is approximately 20 minutes.

\subsubsection{\textbf{Action Transformation: Visual GARAT}}
We use GAIfO adversarial training to optimize the action transformation using a PPO agent with parameters listed in Table. \ref{tab:ppo_hyperparameter}.  The observation space of the agent and discriminator is the concatenation of the policy feature space, $f(o_t)$ (dim = 50) and action space of the environment.  The discriminator learns to differentiate $(f(o_t),a,f(o_{t+1}))$ tuples.  The agent and discriminator are both 2-layer MLPs with hidden dimensions of 1024.  Following \citet{desai2020imitation}, we add the output of the agent to the original action and use action smoothing, proposed in \citet{hanna2017grounded} with smoothing parameter 0.95, to get the transformed action.  Every grounding step, we train the action transformation for 10 epochs. Training time for the action transformation training in each grounding step is approximately 30 minutes.

\begin{table}[ht]
    \caption{SAC hyperparameters.}
    \label{tab:sac_hyperparameter}
    \centering
    \begin{tabular}{lc}
        \toprule
        Hyperparameter & Value \\
        \midrule
        Learning Rate & 0.0003 \\
        Learning Rate Decay & Linear decay \\
        Batch Size & 32 \\
        \# Epochs per Update & 10 \\
        Discount Factor & 0.99 \\
        Entropy Coefficient & 0.001 \\
        Reward Scale & 1  \\
        Normalization & False \\
        \bottomrule
    \end{tabular}
\end{table}

\begin{table}[ht]
    \caption{PPO hyperparameters.}
    \label{tab:ppo_hyperparameter}
    \centering
    \begin{tabular}{lc}
        \toprule
        Hyperparameter & Value \\
        \midrule
        Rollout Size & 5000 \\
        Learning Rate & 0.0003 \\
        Learning Rate Decay & Linear decay \\
        Batch Size & 32 \\
        \# Epochs per Update & 5 \\
        Discount Factor & 0.5 \\
        Entropy Coefficient & 0.001 \\
        Clipping Ratio & 0.1  \\
        Normalization & False  \\
        \bottomrule
    \end{tabular}
\end{table}

\end{document}